\documentclass[sigconf,screen,svgnames,table,hyphens,hidelinks,nonacm]{acmart}

\setcopyright{none}

\usepackage[utf8]{inputenc}
\usepackage[T1]{fontenc}
\usepackage{microtype}
\usepackage{booktabs}
\usepackage{amsmath}
\usepackage{textcomp}
\usepackage{xcolor}
\usepackage{pgfpages}
\usepackage{pgfplots}
\usepackage{pgfplotstable}
\usepackage{tikz}
\usepackage{enumerate}
\usepackage[noend]{algpseudocode}
\usepackage{listings}
\usepackage{tabularx}
\usepackage{url}
\usepackage[multiple]{footmisc}
\usepackage[font=bf]{caption}
\usepackage{multirow}
\usepackage{arydshln}
\usepackage{xspace}
\usepackage{flushend}
\usepackage{diagbox}

\usepackage{bm}
\usepackage{siunitx}
\usepackage{multirow}
\usepackage{enumitem}
\usepackage[linesnumbered,ruled,vlined]{algorithm2e}
\usepackage{graphicx}
\usepackage{subfig}
\usepackage{wrapfig}

\definecolor{light-gray}{gray}{0.75}
\definecolor{BrickRed}{rgb}{0.8, 0.25, 0.33}
\definecolor{lightorange}{HTML}{FFB74D}

\makeatletter
\newenvironment{btHighlight}[1][]
{\begingroup\tikzset{bt@Highlight@par/.style={#1}}\begin{lrbox}{\@tempboxa}}
{\end{lrbox}\bt@HL@box[bt@Highlight@par]{\@tempboxa}\endgroup}

\newcommand\btHL[1][]{%
  \begin{btHighlight}[#1]\bgroup\aftergroup\bt@HL@endenv%
}
\def\bt@HL@endenv{%
  \end{btHighlight}%
  \egroup
}
\newcommand{\bt@HL@box}[2][]{%
  \tikz[#1]{%
    \pgfpathrectangle{\pgfpoint{0.3pt}{0pt}}{\pgfpoint{\wd #2}{\ht #2}}%
    \pgfusepath{use as bounding box}%
    \node[anchor=base west,fill=lightorange,outer sep=0pt,inner xsep=0.3pt,inner ysep=0pt,minimum height=\ht\strutbox+0.3pt,#1]{\raisebox{0.3pt}{\strut}\strut\usebox{#2}};
  }%
}
\makeatother

\usetikzlibrary{calc,trees,positioning,arrows,chains,shapes.geometric,%
  decorations.pathreplacing,decorations.pathmorphing,decorations.text,shapes,%
  matrix,shapes.symbols,patterns,shadows}

\pgfplotsset{width=7cm,compat=newest}
\usepgfplotslibrary{
  colorbrewer,
  groupplots
}

\algrenewcommand\alglinenumber[1]{\tiny\color{Black!70}{#1}}
\algrenewcommand\algorithmicforall[2]{\textbf{for} $i=$ #1 \textbf{to} #2}
\algrenewtext{ForAll}{\algorithmicforall}
\algnewcommand\algorithmicswitch{\textbf{switch}}
\algnewcommand\algorithmiccase{\textbf{case}}
\algdef{SE}[SWITCH]{Switch}{EndSwitch}[1]{\algorithmicswitch\ #1\ \algorithmicdo}{\algorithmicend\ \algorithmicswitch}%
\algdef{SE}[CASE]{Case}{EndCase}[1]{\algorithmiccase\ #1}{\algorithmicend\ \algorithmiccase}%
\algtext*{EndSwitch}%
\algtext*{EndCase}%
\algdef{SE}[SUBALG]{Indent}{EndIndent}{}{\algorithmicend\ }%
\algtext*{Indent}
\algtext*{EndIndent}

\lstdefinestyle{basic}{%
  morekeywords     = [1]{var},%
  morekeywords     = [2]{assert, assume},%
  keywordstyle     = \bfseries\color{DarkBlue},%
  keywordstyle     = [2]\bfseries\color{BrickRed},
  commentstyle     = \ttfamily\color{Black!70}\lst@ifdisplaystyle\footnotesize\fi,%
  basicstyle       = \ttfamily\lst@ifdisplaystyle\footnotesize\fi,%
  columns          = [c]fixed,%
  aboveskip        = 0mm,%
  belowskip        = 2mm,%
  keepspaces       = true,%
  mathescape       = true,%
  escapechar       = ?,%
  tabsize          = 2,%
  numbers          = left,%
  numberstyle      = \tiny\color{Black!80},%
  numbersep        = 1.0em,%
  stepnumber       = 1,%
  firstnumber      = 1,%
  showstringspaces = false,%
  captionpos       = b,%
  extendedchars    = true,%
  xleftmargin      = 2.5em,%
  upquote          = true,%
  abovecaptionskip = 0.5em,%
  belowcaptionskip = 0.5em,%
  moredelim        = **[is][{\btHL[fill=light-gray]}]{°}{°},%
}

\lstdefinestyle{clang}{%
  language         = C,%
  style            = basic,%
}

\newtheoremstyle{mydefinition}
  {}
  {}
  {}
  {}
  {\sc\bfseries}
  {.}
  { }
  {}%

\theoremstyle{mydefinition}
\newtheorem{definition}{Definition}

\newtheorem{theorem}{Theorem}
\newtheorem{corollary}{Corollary}

\newcommand\secref[1]{Sect.~\ref{#1}}

\newcommand\figref[1]{Fig.~\ref{#1}}

\newcommand\algoref[1]{Alg.~\ref{#1}}

\newcommand\thmref[1]{Thm.~\ref{#1}}
\newcommand\corref[1]{Cor.~\ref{#1}}

\newcommand\tool{\textsf{DeepSearch}\xspace}

\usepackage{balance}

\begin{document}

\title[\tool: A Simple and Effective Blackbox Attack for Deep Neural Networks]{\tool: A Simple and Effective Blackbox Attack\\for Deep Neural Networks}

\author{Fuyuan Zhang}
 \affiliation{
   \institution{MPI-SWS \country{Germany}}
 }
\email{fuyuan@mpi-sws.org}

\author{Sankalan Pal Chowdhury}
 \affiliation{
   \institution{MPI-SWS \country{Germany}}
 }
\email{sankalan@mpi-sws.org}

\author{Maria Christakis}
 \affiliation{
   \institution{MPI-SWS \country{Germany}}
 }
\email{maria@mpi-sws.org}

\begin{abstract}
Although deep neural networks have been very successful in image-classification tasks,
they are prone to adversarial attacks. To generate
adversarial inputs, there has emerged a wide variety of techniques,
such as black- and whitebox attacks for neural networks. In this paper,
we present \tool, a novel fuzzing-based, query-efficient, blackbox attack for image
classifiers. Despite its
simplicity, \tool is shown to be more effective in finding adversarial
inputs than state-of-the-art blackbox approaches. \tool is
additionally able to generate the most subtle adversarial inputs in comparison to
these approaches. 
\end{abstract}

\maketitle

\section{Introduction}\label{sec:introduction}
Deep neural networks have been impressively successful in pattern
recognition and image classification \cite{HintonDeng2012,LeCunBottou1998,KrizhevskySutskever2017}. However, it is intriguing that deep neural networks are extremely vulnerable to adversarial attacks \cite{SzegedyZaremba2014}. In fact, even very subtle perturbations of a correctly classified image,
imperceptible to the human eye, may cause a deep neural network to
change its prediction. This poses serious security risks to deploying deep neural networks in safety critical applications. 

Various adversarial attacks have been developed to evaluate the vulnerability of
neural networks against adversarial perturbations.
Early work on generating adversarial examples has focused on whitebox attacks \cite{GoodfellowShlens2015,KurakinGoodfellow2017,MoosaviDezfooliFawzi2016,PapernotMcDaniel2016-Limitations,MadryMakelov2018,CarliniWagner2017-Robustness}. In the whitebox setting, attackers have full access to the network under evaluation, which enables them to calculate gradients of the network. Many gradient-based attacks have been shown to be highly effective. However, in several real-world scenarios, having complete access to network parameters is not realistic. This has motivated the development of blackbox adversarial attacks. 

In the blackbox setting, attackers assume no knowledge about the network structure or its parameters and may only query the target network for its prediction when given particular inputs. One important metric to measure the efficiency of blackbox attacks is the number of queries needed, because queries are essentially time and monetary costs
for attackers, e.g., each query to an online, commercial
machine-learning service costs money. Evaluating the robustness of deep neural networks in a query-limited
blackbox setting is already standard. Gradient-estimation-based blackbox attacks \cite{ChenZhang2017,BhagojiHe2018}, although effective, require a huge number of queries, which makes generating an attack too costly. Various state-of-the-art blackbox
attacks (e.g., \cite{IlyasEngstrom2018,IlyasEngstrom19,GuoGardner19,MoonAn2019}) can already achieve successful attacks with low
number of queries. However, constructing query-efficient blackbox attacks is still open and challenging.

In this paper, we develop a blackbox fuzzing-based technique for
evaluating adversarial robustness of neural networks. The two key challenges of applying fuzzing here
are $(1)$ to maintain a high attack success rate, and $(2)$ to require a low
number of queries. In many cases, without careful guidance while searching, a naive fuzzing
approach, e.g., random fuzzing, is not able find adversarial examples even after a huge number of
queries.  To improve attack success rate, we introduce
carefully designed feedback to guide our search so that images are
efficiently fuzzed toward the decision boundaries. To reduce the number of queries, we adapt
hierarchical grouping~\cite{MoonAn2019} to our setting so that multiple dimensions can
be fuzzed simultaneously, which dramatically reduces query
numbers. Furthermore, a refinement step, which can be viewed as a backward
search step for fuzzing, can effectively reduce distortion of
adversarial examples. Therefore, we extend fuzz testing and show how
to apply it on neural networks in a blackbox setting.

\paragraph{\textbf{Our approach.}}
Inspired by the linear explanation of adversarial
examples \cite{GoodfellowShlens2015}, we develop \tool, a simple, yet
effective, query-efficient, blackbox attack, which is based on
feedback-directed fuzzing. \tool targets deep neural networks for
image classification.
Our attack is constrained by the $L_{\infty}$ distance and only queries the attacked network for its prediction scores on perturbed inputs. The design of our approach is based on the following three aspects:

\begin{enumerate}
	\item \emph{Feedback-directed fuzzing}: Starting from a correctly classified image, \tool strategically mutates its pixels to values that are more likely to lead to an adversarial input. The fuzzing process continues until it either finds an adversarial input or it reaches the query limit. 
	
	\item \emph{Iterative refinement}: Once an adversarial input is found, our approach starts a refinement step to reduce the $L_{\infty}$ distance of this input. The iteration of refinement continues until either the query limit is reached or some termination criterion is met. Our evaluation shows that iterative refinement is able to find subtle adversarial inputs generated by only slightly perturbing pixels in the original image.
	
	\item \emph{Query reduction}: By utilizing the spatial regularities in input images, \tool adapts an existing hierarchical-grouping strategy \cite{MoonAn2019} to our setting and dramatically reduces the number of queries for constructing successful attacks.
	The query-reduction step significantly improves the efficiency of our fuzzing and refinement process.	
\end{enumerate}

We evaluate \tool against four state-of-the-art blackbox attacks in a query-limited setting, where attackers have only a limited query budget to construct attacks. For our evaluation, we use three popular datasets, namely SVHN \cite{NetzerWang11}, CIFAR-$10$~\cite{Krizhevsky2008}, and ImageNet \cite{RussakovskyDeng15}. For SVHN and CIFAR-$10$, we further attack neural networks with state-of-the-art defenses based on adversarial training \cite{MadryMakelov2018}. Our experimental results show that \tool is the most effective in attacking both defended and undefended neural networks. Moreover, it outperforms the other four attacks. Although it is important to develop defense techniques against blackbox adversarial attacks, it is not the focus of this paper and we leave it for future work.

\paragraph{\textbf{Contributions.}}
We make the following contributions:
\begin{enumerate}
	\item We present a simple, yet very effective, fuzzing-based blackbox attack
	for deep neural networks.
	
	\item We perform an extensive evaluation demonstrating that \tool
	is more effective in finding adversarial examples than
	state-of-the-art blackbox approaches.
	
	\item We show that the refinement step in our approach gives \tool the advantage of finding the most subtle
	adversarial examples in comparison to related approaches.
	
	\item We show that the hierarchical-grouping strategy is effective for query reduction in our setting.
	
\end{enumerate}

\paragraph{\textbf{Outline.}}
The next section briefly introduces background. In \secref{sec:deepsearch-binary}, we present \tool for
binary classifiers, that is, networks that classify inputs into two
classes. \secref{sec:deepsearch-multiclass} generalizes the technique
to multiclass classifiers, which classify inputs into multiple
classes. In \secref{sec:deepsearch-with-refinement}, we extend our
technique with iterative refinement such that the generated adversarial examples are
more subtle. We adapt hierarchical grouping for query reduction in \secref{sec:deepsearch-hierarchical-grouping}. We present our experimental evaluation
in \secref{sec:Experiments}, discuss related work
in \secref{sec:RelatedWork}, and conclude in \secref{sec:Conclusion}.
\section{Background}\label{sec:background}
In this section, we introduce some notation and terminology.
Let $\mathbb{R}^{n}$ be the $n$-dimensional vector
space for input images. We represent images as column vectors
$\textbf{x}=(x_{1},...,x_{n})^{T}$, where $x_{i}\in\mathbb{R}$ ($1\leq
i\leq n$) is the $i$th coordinate of $\textbf{x}$. We also write
$\textbf{x}(i)$ to denote the $i$th coordinate $x_{i}$, i.e.,
$\textbf{x}(i)=x_{i}$, and each such coordinate represents an image
pixel.
Now, let $C_{m}=\{l_{1},...,l_{m}\}$ be a set of labels for $m$
classes, where $l_{i}$ is the label for the $i$th class ($1\leq i\leq
m$). A deep neural network that classifies images from
$\mathbb{R}^{n}$ into $m$ classes in $C_{m}$ is essentially a function
$\mathcal{N}: \mathbb{R}^{n}\rightarrow C_{m}$. For an input
$\textbf{x}\in\mathbb{R}^{n}$, $\mathcal{N}(\textbf{x})$ is the label
that the network assigns to $\textbf{x}$.

Assume that input $\textbf{x}$ is correctly classified, and
$\textbf{x}'$ is generated by applying subtle perturbations to
$\textbf{x}$. These perturbations are subtle when the distance between
$\textbf{x}$ and $\textbf{x}'$ in $\mathbb{R}^{n}$ is sufficiently
small according to a distance metric. When this is so and
$\mathcal{N}(\textbf{x})\neq\mathcal{N}(\textbf{x}')$, we say that
$\textbf{x}'$ is an \emph{adversarial example}~\cite{SzegedyZaremba2014}. In
other words, the network is tricked into classifying $\textbf{x}'$
into a different class than $\textbf{x}$ even though they are very
similar.

In this paper, we use the $L_{\infty}$ distance metric. The
$L_{\infty}$ distance between $\textbf{x}$ and $\textbf{x}'$ is
defined as the maximum of their differences along any coordinate
dimension $i$ ($1\leq i\leq n$):
\begin{displaymath}
	||\textbf{x}-\textbf{x}'||_{L_{\infty}} = \max_{i} (|x_{i}-x_{i}'|)
\end{displaymath}
For $d\in\mathbb{R}$, we write $\mathcal{B}(\textbf{x},d)$ to denote
the set of images within distance $d$ from $\textbf{x}$, i.e.,
$\mathcal{B}(\textbf{x},d)=\{\textbf{x}'\ |\
||\textbf{x}-\textbf{x}'||_{L_{\infty}}\leq d\}$, which is an
$n$-dimensional cube. Based on the above, a deep neural network $\mathcal{N}$
is \textit{locally robust} for a correctly classified input
$\textbf{x}$ with respect to distance $d$ if it assigns the same label
to all images in $\mathcal{B}(\textbf{x},d)$.

We mention here that numerous attacks are optimized for one distance metric (e.g.,
\cite{ChenZhang2017,BhagojiHe2018,IlyasEngstrom2018,GuoGardner19,MoonAn2019,GoodfellowShlens2015,
	KurakinGoodfellow2017,MadryMakelov2018,PapernotMcDaniel2016-Limitations,BrendelRauber2018}), just like ours. Although there exist
other distance metrics, e.g., $L_{0}$ and $L_{2}$, and extending attacks from one metric to another is possible,
developing an attack that performs best in all distance metrics is not
realistic. Many state-of-the-art attacks are the most effective in one
metric, but their extension to other metrics performs worse than attacks
specifically designed for that metric. Our paper focuses on a
query-efficient $L_{\infty}$ attack (as in \cite{IlyasEngstrom2018,MoonAn2019}), and our technique
outperforms the state-of-the-art in this setting.

\paragraph{\textbf{Query-limited blackbox threat model.}} We assume that attackers have no knowledge of the target network and can only query the network for its prediction scores, e.g., logits or class probabilities. Moreover, we assume that attackers have a query budget, which can be viewed as time or monetary limits in real-world settings. Thus, the blackbox attack we consider in this paper can be described as follows. Given an input $\textbf{x}$, distance $d$, and query budget $L$, an attacker aims to find an adversarial example $\textbf{x}'$ in  $\mathcal{B}(\textbf{x},d)$ by making at most $L$ queries to the neural network.
\section{Fuzzing Binary Classifiers}\label{sec:deepsearch-binary}
In this section, we present the technical details of how \tool fuzzes
(linear and non-linear) binary classifiers. We first introduce our approach for linear binary classifiers, which serves as the
mathematical foundation. Then, we generalize our approach to non-linear
binary classifiers through iterative linear approximations.

\subsection{Linear Binary Classifiers}\label{subsec:linear-binary-classifiers}

A \textit{binary classifier} classifies inputs into two classes,
denoted with labels $C_{2}=\{l_{1},l_{2}\}$, according to the
definition below.
\begin{definition}[{\bf Binary Classifier}]
	Given a classification function
	$f:\mathbb{R}^{n}\rightarrow\mathbb{R}$, a \textit{binary classifier}
	$\mathcal{N}_{f}:\mathbb{R}^{n}\rightarrow C_{2}$ is defined as
	follows:
	\begin{displaymath}
		\mathcal{N}_{f}(\textbf{x}) = \left\{ \begin{array}{ll}
			l_{1}, & \hspace{0.5cm}\textrm{if $f(\textbf{x})>0$}\\
			l_{2}, & \hspace{0.5cm}\textrm{if $f(\textbf{x})<0$}
		\end{array} \right.
	\end{displaymath}
	If function $f$ is linear, then $\mathcal{N}_{f}$ is a \emph{linear}
	binary classifier, otherwise it is non-linear.
\end{definition}

The set of values $\mathcal{D}_{f}=\{\textbf{x}\ |\ f(\textbf{x})=0\}$
constitute the \textit{decision boundary} of $\mathcal{N}_{f}$, which
classifies the domain $\mathbb{R}^{n}$ into the two classes in
$C_{2}$.

\begin{figure}[t!]
	\centering
	\vspace{-15pt}
	\subfloat[Linear classifier]{{\includegraphics[width=3.8cm]{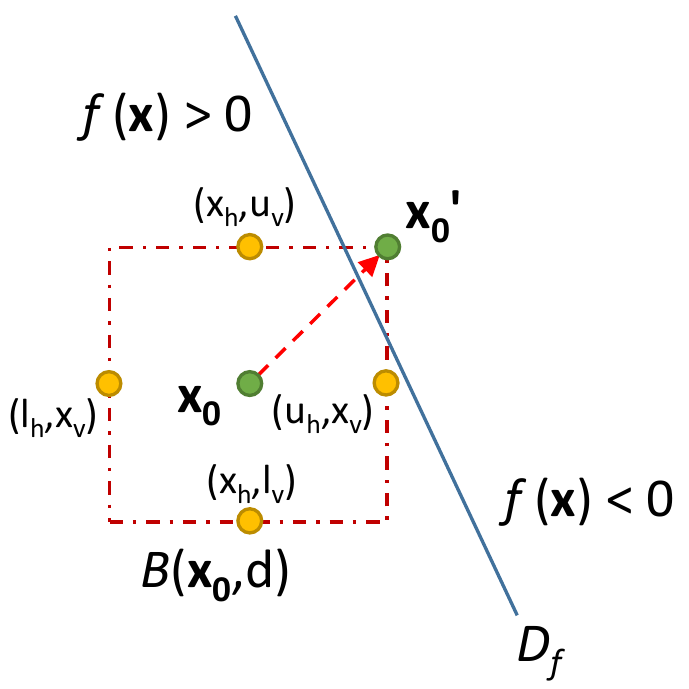}}\label{fig-deep-search-binary-linear}}%
	\vspace{-5pt}
	\qquad
	\subfloat[Non-linear classifier]{{\includegraphics[width=3.8cm]{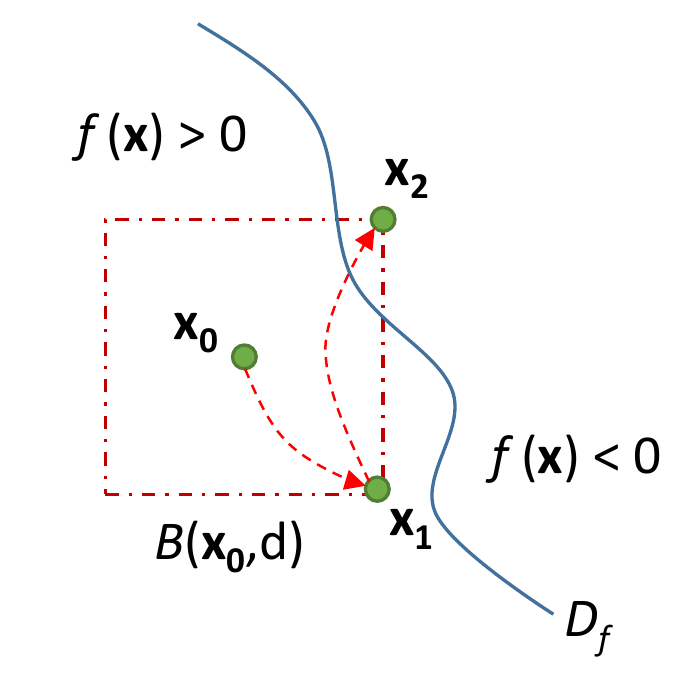}}\label{fig-deep-search-binary-nonlinear}}%
	\caption{\tool for binary classifiers.}
	\label{fig-deep-search-binary}
	\vspace{-1em}
\end{figure}

As an example, consider \figref{fig-deep-search-binary-linear},
showing a linear binary classifier
$\mathcal{N}_{f}:\mathbb{R}^{2}\rightarrow C_{2}$. Observe that input
$\textbf{x}_{0}$ is classified in $l_1$ whereas $\textbf{x}'_{0}$ is in
$l_2$. Note that the decision boundary of a linear classifier
$\mathcal{N}_{f}:\mathbb{R}^{n}\rightarrow C_{2}$ is a hyperplane; it
is, therefore, a straight line in
\figref{fig-deep-search-binary-linear}.
Now, assume that $\textbf{x}_{0}$ is correctly classified and that the
dash-dotted square represents $\mathcal{B}(\textbf{x}_{0},d)$. Then,
$\textbf{x}'_{0}$ is adversarial because $\mathcal{N}_f(\textbf{x}_{0})\neq\mathcal{N}_f(\textbf{x}'_{0})$,
which is equivalent to $f(\textbf{x}_{0})f(\textbf{x}'_{0})<0$.

\paragraph{\textbf{Example.}}
We give an intuition on how \tool handles linear binary classifiers
using the example of \figref{fig-deep-search-binary-linear}. Recall
that $\textbf{x}_{0}$ is a correctly classified input for which
$f(\textbf{x}_{0})>0$. To find an adversarial example, \tool fuzzes
$\textbf{x}_{0}$ with the goal of generating a new input
$\textbf{x}'_{0}$ such that $f(\textbf{x}'_{0})<0$.

Fuzzing is performed as follows. Input $\textbf{x}_{0}$ has two
coordinates $x_h$ and $x_v$, for the horizontal and vertical
dimensions. \tool independently mutates each of these coordinates to
the minimum and maximum values that are possible within
$\mathcal{B}(\textbf{x}_{0},d)$, with the intention of finding the minimum value 
of $f$ in $\mathcal{B}(\textbf{x}_{0},d)$. 
For instance, when mutating $x_h$, we
obtain inputs $\textbf{x}_{0}[l_h/x_h]$ and $\textbf{x}_{0}[u_h/x_h]$
in the figure. Values $l_h$ and $u_h$ are, respectively, the minimum
and maximum that $x_h$ may take, and $\textbf{x}_{0}[l_h/x_h]$ denotes
substituting $x_h$ with $l_h$ (similarly for
$\textbf{x}_{0}[u_h/x_h]$).
We then evaluate $f(\textbf{x}_{0}[l_h/x_h])$ and
$f(\textbf{x}_{0}[u_h/x_h])$, and for $x_h$, we select the value
($l_h$ or $u_h$) that causes function $f$ to \emph{decrease}. This is
because, in our example, an adversarial input $\textbf{x}'_{0}$ must
make the value of $f$ negative. Let us assume that
$f(\textbf{x}_{0}[u_h/x_h]) < f(\textbf{x}_{0}[l_h/x_h)$; we, thus,
select $u_h$ for coordinate $x_h$.

\tool mutates coordinate $x_v$ in a similar way. It evaluates
$f(\textbf{x}_{0}[l_v/x_v])$ and $f(\textbf{x}_{0}[u_v/x_v])$, and
selects the value that causes $f$ to decrease. Let us assume that
$f(\textbf{x}_{0}[u_v/x_v]) < f(\textbf{x}_{0}[l_v/x_v])$; we, thus,
select $u_v$ for $x_v$.

Next, we generate input $\textbf{x}'_{0}$ by substituting each
coordinate in $\textbf{x}_{0}$ with the boundary value that was
previously selected. In other words, $\textbf{x}'_{0} =
\textbf{x}_{0}[u_h/x_h,u_v/x_v]$, and since $f(\textbf{x}'_{0}) < 0$,
\tool has generated an adversarial example. Note that $f(\textbf{x}'_{0})$ is actually
the minimum value of $f$ in $\mathcal{B}(\textbf{x}_{0},d)$.

\paragraph{\textbf{\tool for linear binary classifiers.}}
We now formalize how \tool treats linear binary classifiers.
Consider a linear classification function
$f(\textbf{x})=\textbf{w}^{T}\textbf{x}+b=\sum_{i=1}^{n}w_{i}x_{i}+b$,
where $\textbf{w}^{T}=(w_{1},...,w_{n})$ and $b\in\mathbb{R}$. Note
that $f$ is monotonic with respect to all of its variables
$x_{1},...,x_{n}$. For instance, if $w_{i} > 0$, then $f$ is
monotonically increasing in $x_{i}$.

Recall that $\mathcal{B}(\textbf{x},d)$ denotes the set of inputs
within distance $d\in\mathbb{R}$ of an input
$\textbf{x}$. $\mathcal{B}(\textbf{x},d)$ may be represented by an
$n$-dimensional cube $\mathcal{I}=I_{1}\times...\times I_{n}$, where
$I_{i}=[l_{i},u_{i}]$ is a closed interval bounded by
$l_{i},u_{i}\in\mathbb{R}$ with $l_{i}\leq u_{i}$ for $1\leq i\leq
n$. Intuitively, value $l_{i}$ (resp. $u_{i}$) is the lower
(resp. upper) bound on the $i$th dimension of $\textbf{x}$. An input
$\textbf{x}'$ is a \textit{vertex} of $\mathcal{I}$ if each of its
coordinates $\textbf{x}'(i)$ is an endpoint of $I_{i}$ for $1\leq
i\leq n$, i.e., $\textbf{x}'(i)=u_{i}$ or $l_{i}$ ($1\leq i\leq n$).

Due to the monotonicity of $f$, the maximum and minimum values of $f$
on $\mathcal{I}$ may be easily calculated by applying $f$ to vertices
of $\mathcal{I}$. For example, consider a one-dimensional linear function $f(x) = -2x$, where $x
\in [-1, 1]$, that is, $-1$ and $1$ are the lower and upper bounds for
$x$. Since $f(1) < f(-1)$, we get a maximum value of $f$ at $x = -1$
and a minimum value of $f$ at $x = 1$. $N$-dimensional linear functions
can be treated similarly. We write $f(\mathcal{I})$ for the values of $f$ on $\mathcal{I}$, i.e., 
$f(\mathcal{I})=\{f(\textbf{x})\ |\ \textbf{x}\in \mathcal{I}\}$, and have the following theorem (whose proof can be found in \cite{ZhangChowdhury2020}).

\begin{theorem}\label{theorem-maximum-minimum}
	Given a linear classification function
	$f(\textbf{x})=\textbf{w}^{T}\textbf{x}+b$, where
	$\textbf{w}^{T}=(w_{1},...,w_{n})$ and $b\in\mathbb{R}$, an
	$n$-dimensional cube $\mathcal{I}=I_{1}\times...\times I_{n}$, where
	$I_{i}=[l_{i},u_{i}]$ for $1\leq i\leq n$, and an input
	$\textbf{x}\in \mathcal{I}$, we have:
	\begin{enumerate}
		\item $\min$ $f(\mathcal{I}) = f(\textbf{x}')$, where
		$\textbf{x}'(i)=l_{i}$ (resp. $\textbf{x}'(i)=u_{i}$) if
		$f(\textbf{x}[u_{i}/x_{i}])> f(\textbf{x}[l_{i}/x_{i}])$
		(resp. $f(\textbf{x}[u_{i}/x_{i}])\leq
		f(\textbf{x}[l_{i}/x_{i}])$) for $1\leq i\leq
		n$
		\item $\max$ $f(\mathcal{I}) =
		f(\textbf{x}')$, where $\textbf{x}'(i)=u_{i}$
		(resp. $\textbf{x}'(i)=l_{i}$) if $f(\textbf{x}[u_{i}/x_{i}])>
		f(\textbf{x}[l_{i}/x_{i}])$
		(resp. $f(\textbf{x}[u_{i}/x_{i}])\leq
		f(\textbf{x}[l_{i}/x_{i}])$) for $1\leq i\leq n$
	\end{enumerate}
\end{theorem}

According to the above theorem, we can \emph{precisely} calculate the
minimum and maximum values of $f$ in any $n$-dimensional cube. In
particular, assume a correctly classified input $\textbf{x}$ for which
$f(\textbf{x})>0$. For each dimension $i$ ($1\leq i\leq n$) of
$\textbf{x}$, we first construct inputs $\textbf{x}[l_{i}/x_{i}]$ and
$\textbf{x}[u_{i}/x_{i}]$. We then compare the values of
$f(\textbf{x}[l_{i}/x_{i}])$ and $f(\textbf{x}[u_{i}/x_{i}])$. To
generate a new input $\textbf{x}'$, we select the value of its $i$th
coordinate as follows:
\begin{displaymath}
	\textbf{x}'(i) = \left\{ \begin{array}{ll}
		l_{i}, & \hspace{0.5cm}\textrm{if $f(\textbf{x}[u_{i}/x_{i}])> f(\textbf{x}[l_{i}/x_{i}])$}\\
		u_{i}, & \hspace{0.5cm}\textrm{if $f(\textbf{x}[u_{i}/x_{i}])\leq f(\textbf{x}[l_{i}/x_{i}])$}
	\end{array} \right.
\end{displaymath}

As shown here, selecting a value for $\textbf{x}'(i)$ requires
evaluating function $f$ twice, i.e., $f(\textbf{x}[l_{i}/x_{i}])$ and
$f(\textbf{x}[u_{i}/x_{i}])$. Therefore, for $n$ dimensions, $f$ must
be evaluated $2n$ times.
In practice however, due to the monotonicity of $f$, evaluating it
only once per dimension is sufficient. For instance, if $f(\textbf{x}[u_{i}/x_{i}])$
already decreases (resp. increases) the value of $f$ in comparison to
$f(\textbf{x})$, there is no need to evaluate
$f(\textbf{x}[l_{i}/x_{i}])$. Value $u_{i}$ (resp. $l_{i}$) should be
selected for the $i$th coordinate. Hence, the minimum value of $f$ can
be computed by evaluating the function exactly $n$ times.
If, for the newly generated input $\textbf{x}'$, the sign of $f$
becomes negative, $\textbf{x}'$ constitutes an adversarial example.

We treat the case where $f(\textbf{x})<0$ for a correctly classified
input $\textbf{x}$ analogously. \tool aims to generate a new input
$\textbf{x}'$ such that the sign of $f$ becomes positive. We are,
therefore, selecting coordinate values that cause $f$
to \emph{increase}.

\begin{algorithm}[t!]
	\small
	\DontPrintSemicolon
	\SetKwProg{Fn}{Function}{is}{end}
	\SetKwFunction{Max}{ApproxMax}
	\SetKwFunction{Min}{ApproxMin}
	\SetKwFunction{DeepSearchBinary}{DS-Binary}
	\KwIn{input $\textbf{x}\in\mathbb{R}^{n}$, initial input $\textbf{x}_{\mathit{init}}\in\mathcal{B}(\textbf{x},d)$,\\\hspace{3em}function $f:\mathbb{R}^{n}\rightarrow \mathbb{R}$, distance $d\in\mathbb{R}$}
	\KwOut{$\textbf{x}'\in\mathcal{B}(\textbf{x},d)$}
	
	\Fn{\Max{\normalfont{$\textbf{x}, f, (I_{1},...,I_{n})$}} \ }{
		$\textbf{x}':= (0,...,0)$\; \ForEach{$1\leq i\leq
			n$}{ \eIf{\normalfont{$f(\textbf{x}[u_{i}/x_{i}])>
					f(\textbf{x}[l_{i}/x_{i}])$}}{
				$\textbf{x}':= \textbf{x}'[u_{i}/x_{i}']$ }{
				$\textbf{x}':= \textbf{x}'[l_{i}/x_{i}']$ } } \Return{\normalfont{$\textbf{x}'$}}
	}
	\;
	\Fn{\Min{\normalfont{$\textbf{x}, f, (I_{1},...,I_{n})$}} \ }{
		$\textbf{x}':= (0,...,0)$\;
		\ForEach{$1\leq i\leq n$}{
			\eIf{\normalfont{$f(\textbf{x}[u_{i}/x_{i}])>f(\textbf{x}[l_{i}/x_{i}])$}}{
				$\textbf{x}':= \textbf{x}'[l_{i}/x_{i}']$
			}{
				$\textbf{x}':= \textbf{x}'[u_{i}/x_{i}']$
			}
		}
		\Return{\normalfont{$\textbf{x}'$}}
		
	}
	\;
	\Fn{\DeepSearchBinary{\normalfont{$\textbf{x},\textbf{x}_{\mathit{init}}, f, d$}} \ }{
		construct intervals $(I_{1},...,I_{n})$ such that $\mathcal{B}(\textbf{x},d)=I_{1}\times...\times I_{n}$\;
		initialize $\textbf{x}_{0}:=\textbf{x}_{\mathit{init}}$ and $k:= 0$\;
		\eIf {\normalfont{$f(\textbf{x}_{0})>0$}}{
			\Repeat {$\mathcal{N}_{f}(\textbf{x})\neq\mathcal{N}_{f}(\textbf{x}_{k})$, or $k=\texttt{MaxNum}$}{
				$\textbf{x}_{k+1}:= \Min\ (\textbf{x}_{k}, f, (I_{1},...,I_{n}))$\;
				$k:=k+1$
			}
		}{
			\Repeat {$\mathcal{N}_{f}(\textbf{x})\neq\mathcal{N}_{f}(\textbf{x}_{k})$, or $k=\texttt{MaxNum}$}{
				$\textbf{x}_{k+1}:=\Max\ (\textbf{x}_{k}, f, (I_{1},...,I_{n}))$\;
				$k:=k+1$
				
			}
		}
		
		\Return {\normalfont{$\textbf{x}_{k}$}}\;
		
	}
	\caption{\textbf{\tool for binary classifiers.}}
	\label{alg:algorithm-binary-classifiers}
\end{algorithm}

\subsection{Non-Linear Binary Classifiers}\label{subsec:nonlinear-binary-classifiers}

We generalize our technique to non-linear binary classifiers. In this
setting, \tool iteratively \emph{approximates} the minimum and maximum
values of $f$ in $\mathcal{B}(\textbf{x},d)$. 

\paragraph{\textbf{Example.}}
As an example, consider the non-linear classification function $f$
shown in \figref{fig-deep-search-binary-nonlinear}. Since $f$ is
non-linear, the decision boundary $\mathcal{D}_{f}$ of the binary
classifier is a curve.

Starting from correctly classified input $\textbf{x}_{0}$, \tool
treats $f$ as linear within $\mathcal{B}(\textbf{x}_{0},d)$ and
generates $\textbf{x}_{1}$, exactly as it would for a linear binary
classifier. To explain how $\textbf{x}_{1}$ is derived, we refer to the points in \figref{fig-deep-search-binary-linear}.
Suppose we first mutate the horizontal dimension of $\textbf{x}_{0}$ (using
the lower and upper bounds) and find that $f(l_h, x_v) > f(u_h,
x_v)$. To increase the chances of crossing the decision boundary, we
choose the bound for the horizontal dimension of $\textbf{x}_{0}$ that
gives us the lower value of $f$, i.e., we select $u_h$ for horizontal
coordinate $x_h$. Then, we mutate the vertical dimension of $\textbf{x}_{0}$ and
find that $f(x_h, u_v) > f(x_h, l_v)$. This means that we select $l_v$
for vertical coordinate $x_v$. Hence, we derive $\textbf{x}_{1}=(u_h, l_v)$. 
Observe that input $\textbf{x}_{1}$ is not
adversarial. Unlike for a linear binary classifier however, where the
minimum value of $f$ in $\mathcal{B}(\textbf{x}_{0},d)$ is precisely
computed, the non-linear case is handled by iteratively approximating
the minimum. In particular, after generating $\textbf{x}_{1}$, \tool
iterates starting from $\textbf{x}_{1}$, while again treating $f$ as
linear in $\mathcal{B}(\textbf{x}_{0},d)$. As a result, our technique
generates input $\textbf{x}_{2}$, which is adversarial.

The reason we can treat non-linear binary classifiers as linear ones is that
perturbations of pixels are only allowed in a very small $n$-dimensional cube,
constrained by the $L_{\infty}$ distance. Within such a small space, we
can effectively approximate non-linear functions using iterative
linear approximations.

\paragraph{\textbf{\tool for non-linear binary classifiers.}}
\algoref{alg:algorithm-binary-classifiers} shows \tool for binary
classifiers. It uses iterative approximations to search for
adversarial examples. Note that our technique is blackbox, and
consequently, it cannot differentiate between linear and non-linear
classifiers. \algoref{alg:algorithm-binary-classifiers} is, therefore,
the general algorithm that \tool applies to fuzz any binary
classifier.

The main function in \algoref{alg:algorithm-binary-classifiers} is
\texttt{DS-Binary}. Input $\textbf{x}_{\mathit{init}}$ is the input from
which we start the first iteration, e.g., it corresponds to
$\textbf{x}_{0}$ in \figref{fig-deep-search-binary}. Input
$\textbf{x}$ is used to compute $\mathcal{B}(\textbf{x},d)$, and for
now, assume that $\textbf{x}$ is equal to
$\textbf{x}_{\mathit{init}}$.  (We will discuss why $\textbf{x}$ is
needed in \secref{sec:deepsearch-with-refinement}.) In addition to
these inputs, the algorithm also takes a classification function $f$
and the distance $d$. 

Function \texttt{DS-Binary} assigns $\textbf{x}_{\mathit{init}}$
to $\textbf{x}_{0}$ and constructs $n$ intervals $I_{1},..., I_{n}$ to
represent $\mathcal{B}(\textbf{x}_{0},d)$ (lines 20--21). Then, based
on the sign of $f(\textbf{x}_{0})$, our algorithm iteratively
approximates the minimum (lines 23--26) or the maximum (lines 28--31)
value of $f$ in $\mathcal{B}(\textbf{x}_{0},d)$. 
\texttt{DS-Binary}
terminates when either an adversarial example is found or it has reached \texttt{MaxNum} iterations. To find adversarial examples
in $k$ iterations, we evaluate $f$ at most $2n+n(k-1)$ times. 

\texttt{ApproxMin} and \texttt{ApproxMax}
implement \thmref{theorem-maximum-minimum} to calculate the minimum
and maximum values of function $f$ in the $n$-dimensional cube
$I_{1}\times...\times I_{n}$. When $f$ is linear, calling these
functions on any input $\textbf{x}\in I_{1}\times...\times I_{n}$ does
not affect the computation. In other words, the minimum and maximum
values are precisely computed for any $\textbf{x}$.
When $f$ is non-linear, it is still assumed to be linear within the
$n$-dimensional cube. Given that the size of the cube is designed to
be small, this assumption does not introduce too much imprecision. As
a consequence of this assumption however, different inputs in the
$n$-dimensional cube lead to computing different minimum and maximum
values of $f$. For instance,
in \figref{fig-deep-search-binary-nonlinear},
calling \texttt{ApproxMin} on $\textbf{x}_{0}$ returns
$\textbf{x}_{1}$, while calling it on $\textbf{x}_{1}$ returns
$\textbf{x}_{2}$.
\section{Fuzzing Multiclass Classifiers}\label{sec:deepsearch-multiclass}
In this section, we extend our technique for blackbox fuzzing of
binary classifiers to multiclass classifiers.

\subsection{Linear Multiclass Classifiers}

A \emph{multiclass classifier} classifies inputs in $m$ classes
according to the following definition.

\begin{definition}[{\bf Multiclass Classifier}]
	For classification function $f:\mathbb{R}^{n}\rightarrow
	\mathbb{R}^{m}$, which returns $m$ values each corresponding to one
	class in $C_{m} = \{l_1,...,l_m\}$, a \emph{multiclass classifier}
	$\mathcal{N}_{f}:\mathbb{R}^{n}\rightarrow C_{m}$ is defined as
	\begin{displaymath}
		\begin{array}{rl}
			\mathcal{N}_{f}(\textbf{x}) = l_{j}, & \hspace{0.5cm} \textrm{iff $j = \arg\max_{i} f_{i}(\textbf{x})$},
		\end{array}
	\end{displaymath}
	where $f_{i}:\mathbb{R}^{n}\rightarrow \mathbb{R}$ denotes the
	function derived by evaluating $f$ for the $i$th class, i.e.,
	$f(\textbf{x})=(f_{1}(\textbf{x}),...,f_{m}(\textbf{x}))^{T}$.
\end{definition}

In other words, a multiclass classifier $\mathcal{N}_{f}$ classifies
an input $\textbf{x}$ in $l_{j}$ if $f_{j}(\textbf{x})$ evaluates to
the largest value in comparison to all other functions $f_i$.

\begin{figure}[t!]
	\centering
	\vspace{-15pt}
	\subfloat[Linear classifier]{{\includegraphics[width=3.8cm]{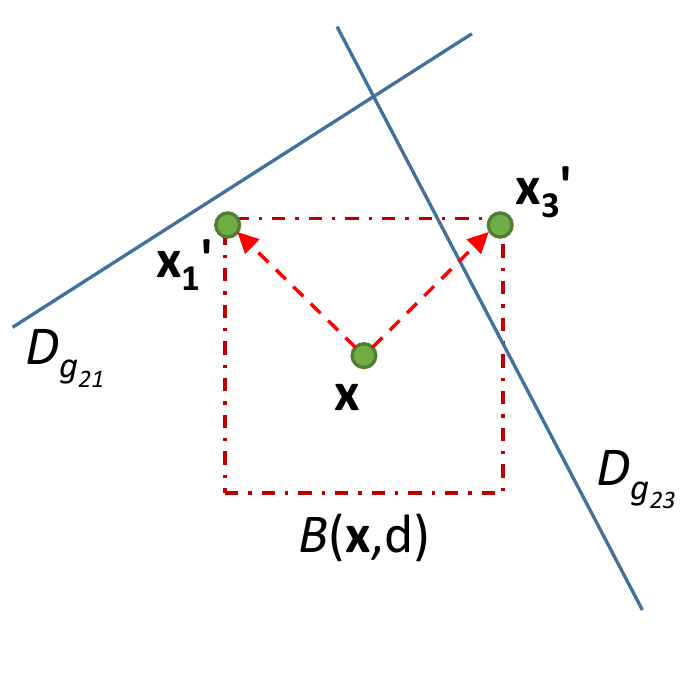}}\label{fig-deep-search-multiclass-linear}}%
	\vspace{-5pt}
	\qquad
	\subfloat[Non-linear classifier]{{\includegraphics[width=3.8cm]{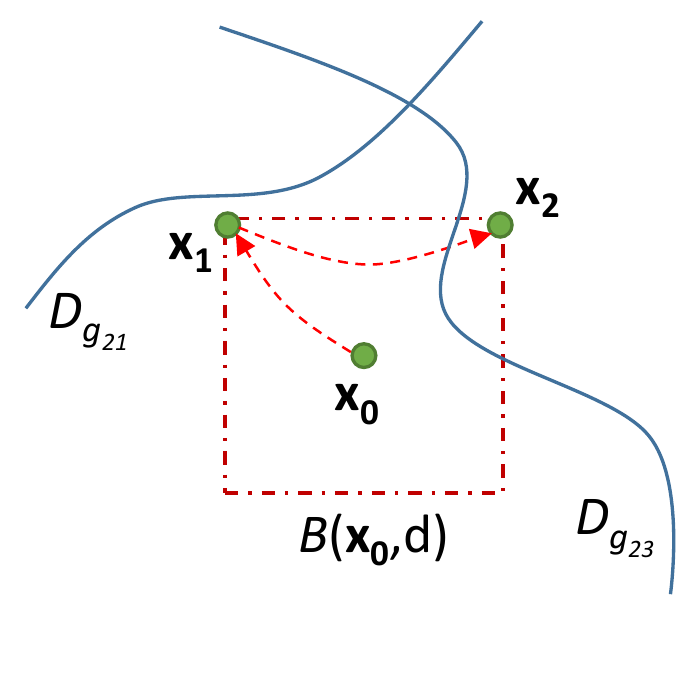}}\label{fig-deep-search-multiclass-nonlinear}}%
	\caption{\tool for multiclass classifiers.}
	\label{fig-deep-search-multiclass}
	\vspace{-1em}
\end{figure}

Function $f$ of a multiclass classifier $\mathcal{N}_{f}$ may be
decomposed into multiple binary classifiers such that the original
classifier can be reconstructed from the binary
ones.
First, to decompose a multiclass classifier into binary classifiers,
for any pair of classes $l_{i}$ and $l_{j}$ $(1\leq i,j\leq m)$, we
define a classification function $g_{ij}:\mathbb{R}^{n}\rightarrow
\mathbb{R}$ as $g_{ij}(\textbf{x})= f_{i}(\textbf{x}) -
f_{j}(\textbf{x})$. We then construct a binary classifier
$\mathcal{N}_{g_{ij}}:\mathbb{R}^{n}\rightarrow \{l_{i},l_{j}\}$ as
follows:
\begin{displaymath}
	\mathcal{N}_{g_{ij}}(\textbf{x}) = \left\{ \begin{array}{ll}
		l_{i}, & \hspace{0.5cm}\textrm{if $g_{ij}(\textbf{x})>0$}\\
		l_{j}, & \hspace{0.5cm}\textrm{if $g_{ij}(\textbf{x})<0$}
	\end{array} \right.
\end{displaymath}
As usual, the set of values
$\mathcal{D}_{g_{ij}}=\{\textbf{x}\ |\ f_{i}(\textbf{x}) -
f_{j}(\textbf{x})=0\}$ constitutes the pairwise decision boundary of
binary classifier $\mathcal{N}_{g_{ij}}$, which classifies the domain
$\mathbb{R}^{n}$ into the two classes $\{l_{i},l_{j}\}$.
As an example, consider \figref{fig-deep-search-multiclass-linear}
depicting a multiclass classifier $\mathcal{N}_{f} :
\mathbb{R}^{2}\rightarrow C_{3}$, where $f$ is linear and
$C_{3}=\{l_{1},l_{2},l_{3}\}$. Assume that $\mathcal{N}_{f}$ correctly
classifies input $\textbf{x}$ in $l_2$. Based on the above, linear
binary classifiers $\mathcal{N}_{g_{21}}$ and $\mathcal{N}_{g_{23}}$
also classify $\textbf{x}$ in $l_2$, i.e., $g_{21}(\textbf{x})>0$ and
$g_{23}(\textbf{x})>0$.

Second, a multiclass classifier may be composed from multiple binary
classifiers as follows. An input $\textbf{x}$ is classified in class
$l_{i}$ by multiclass classifier $\mathcal{N}_{f}$ if and only if it
is classified in $l_{i}$ by all $m-1$ binary classifiers
$\mathcal{N}_{g_{ij}}$ for $1\leq j\leq m, i\neq j$, where $l_{i}\in
C_{m}$ and $g_{ij}(\textbf{x})= f_{i}(\textbf{x}) -
f_{j}(\textbf{x})$:
\begin{displaymath}
	\begin{array}{rl}
		\mathcal{N}_{f}(\textbf{x}) = l_{i}, & \hspace{0.5cm} \textrm{iff $\forall 1\leq j\leq m,i\neq j: \mathcal{N}_{g_{ij}}(\textbf{x})=l_{i}$}
	\end{array}
\end{displaymath}
For instance, in \figref{fig-deep-search-multiclass-linear}, if both
$\mathcal{N}_{g_{21}}$ and $\mathcal{N}_{g_{23}}$ classify input
$\textbf{x}$ in class $l_2$, then the multiclass classifier also
classifies it in $l_2$.

Based on the above, a multiclass classifier has an adversarial input
if and only if this input is also adversarial for a constituent binary
classifier.

\begin{corollary}\label{corollary-multiclass-classifier}
	Let $\mathcal{N}_f$ be a multiclass classifier and
	$\textbf{x}\in\mathbb{R}^{n}$ a correctly classified input, where
	$\mathcal{N}_{f}(\textbf{x})=l_{i}$ and $l_{i}\in C_{m}$. There exists
	an adversarial example $\textbf{x}' \in \mathcal{B}(\textbf{x},d)$ for
	$\mathcal{N}_f$, where $d\in\mathbb{R}$, if and only if $\textbf{x}'$
	is an adversarial example for a binary classifier
	$\mathcal{N}_{g_{ij}}$ ($1\leq j\leq m, i\neq j$), where
	$g_{ij}(\textbf{x})= f_{i}(\textbf{x}) - f_{j}(\textbf{x})$:
	\begin{displaymath}
		\begin{array}{rl}
			\mathcal{N}_{f}(\textbf{x}') \neq l_{i}, & \hspace{0.5cm} \textrm{iff $\exists 1\leq j\leq m, i\neq j: \mathcal{N}_{g_{ij}}(\textbf{x}')\neq l_{i}$}
		\end{array}
	\end{displaymath}
\end{corollary}

\paragraph{\textbf{Example.}}
This corollary is crucial in generalizing our technique to multiclass
classifiers. Assume a correctly classified input $\textbf{x}$, for
which $\mathcal{N}_{f}(\textbf{x})=l_{i}$. According to the above
corollary, the robustness of $\mathcal{N}_{f}$ in
$\mathcal{B}(\textbf{x},d)$ reduces to the robustness of all $m-1$
binary classifiers $\{\mathcal{N}_{g_{ij}}\ |\ 1\leq j\leq m, i\neq
j\}$ in $\mathcal{B}(\textbf{x},d)$. We, therefore, use \tool for
binary classifiers to test each binary classifier in this set. If
there exists an adversarial input $\textbf{x}'$ for one of these
classifiers, i.e., $\mathcal{N}_{g_{ij}}(\textbf{x}') \neq l_i$ for
some $j$, then $\textbf{x}'$ is also an adversarial input for
$\mathcal{N}_{f}$, i.e., $\mathcal{N}_{f}(\textbf{x}') \neq l_i$.

Let us consider again the example of
\figref{fig-deep-search-multiclass-linear}. Recall that multiclass
classifier $\mathcal{N}_{f}$ correctly classifies input $\textbf{x}$
in class $l_2$, and so do binary classifiers $\mathcal{N}_{g_{21}}$
and $\mathcal{N}_{g_{23}}$, i.e., $g_{21}(\textbf{x})>0$ and
$g_{23}(\textbf{x})>0$. As a result, \tool tries to generate inputs
that decrease the value of each of these functions in
$\mathcal{B}(\textbf{x},d)$ in order to find adversarial examples.
Function $g_{21}$ evaluates to its minimum value in
$\mathcal{B}(\textbf{x},d)$ for input $\textbf{x}_{1}'$, and function
$g_{23}$ for input $\textbf{x}_{3}'$. Observe that $\textbf{x}_{3}'$
is an adversarial example for $\mathcal{N}_{g_{23}}$, and thus also
for $\mathcal{N}_{f}$, whereas $\textbf{x}_{1}'$ is not.

\paragraph{\textbf{\tool for linear multiclass classifiers.}}
Let us assume a linear classification function
$f(\textbf{x})=\textbf{W}^{T}\textbf{x}+\textbf{b}$, where
$\textbf{W}^{T}=(\textbf{w}^{T}_{1},...,\textbf{w}^{T}_{m})^{T}$,
$\textbf{w}_{i}\in\mathbb{R}^{n}$ $(1\leq i\leq m)$, and
$\textbf{b}=(b_{1},...,b_{m})^{T}\in\mathbb{R}^{m}$. Then, $f_{i}$,
which denotes the function derived by evaluating $f$ for the $i$th
class, is of the form
$f_{i}(\textbf{x})=\textbf{w}^{T}_{i}\textbf{x}+b_{i}$ for $1\leq
i\leq m$. For any pair of class labels $l_{i}$ and $l_{j}$, function
$g_{ij}$ is defined as $g_{ij}(\textbf{x})= f_{i}(\textbf{x}) -
f_{j}(\textbf{x})=(\textbf{w}^{T}_{i}-\textbf{w}^{T}_{j})\textbf{x}+(b_{i}-b_{j})$.
Hence, $g_{ij}$ is also linear, and $\mathcal{N}_{g_{ij}}$ is a linear
binary classifier.

Assume that classifier $\mathcal{N}_{f}$ correctly classifies input
$\textbf{x}\in\mathbb{R}^{n}$ in $l_i$,
$\mathcal{N}_{f}(\textbf{x})=l_{i}$ $(l_{i}\in C_{m})$. According to
\corref{corollary-multiclass-classifier}, the robustness of
$\mathcal{N}_{f}$ in $\mathcal{B}(\textbf{x},d)$ $(d\in\mathbb{R})$
reduces to the robustness of each binary classifier
$\mathcal{N}_{g_{ij}}$ $(1\leq j\leq m, i\neq j)$ in
$\mathcal{B}(\textbf{x},d)$.
To find an adversarial example for a binary classifier
$\mathcal{N}_{g_{ij}}$ in $\mathcal{B}(\textbf{x},d)$, \tool must
generate an input $\textbf{x}'\in\mathcal{B}(\textbf{x},d)$ such that
$g_{ij}(\textbf{x}')<0$. (Recall that by definition
$g_{ij}(\textbf{x})>0$.)
Since all functions $g_{ij}$ $(1\leq j\leq m, i\neq j)$ are linear, we
easily find their minimum values in $\mathcal{B}(\textbf{x},d)$ as
follows.

Let $I_{1},..., I_{n}$ be intervals such that
$\mathcal{B}(\textbf{x},d)=I_{1}\times...\times I_{n}$, where
$I_{k}=[l_{k},u_{k}]$ for $1\leq k\leq n$. As in
\secref{subsec:linear-binary-classifiers}, for each dimension $k$,
\tool evaluates function $f$ twice to compare the values of
$f(\textbf{x}[u_{k}/x_{k}])$ and $f(\textbf{x}[l_{k}/x_{k}])$. To
generate a new input $\textbf{x}_{j}'$ for which function $g_{ij}$
evaluates to its minimum value, we select its $k$th coordinate as
follows:
\begin{displaymath}
	\textbf{x}_{j}'(k) = \left\{ \begin{array}{ll}
		l_{k}, & \hspace{0.5cm}\textrm{if $g_{ij}(\textbf{x}[u_{k}/x_{k}])> g_{ij}(\textbf{x}[l_{k}/x_{k}])$}\\
		u_{k}, & \hspace{0.5cm}\textrm{if $g_{ij}(\textbf{x}[u_{k}/x_{k}])\leq g_{ij}(\textbf{x}[l_{k}/x_{k}])$.}
	\end{array} \right.
\end{displaymath}

Note that, although we calculate the minimum value of $m-1$ linear
functions, we still evaluate $f$ $2n$ times. This is because a
function $g_{ij}$ is defined as $g_{ij}(\textbf{x})= f_{i}(\textbf{x})
- f_{j}(\textbf{x})$, where $f_i(\textbf{x})$ and $f_j(\textbf{x})$
are the values of $f(\textbf{x})$ for the $i$th and $j$th classes,
respectively. If the sign of $g_{ij}(\textbf{x}_{j}')$ becomes
negative for some $j$, then \tool has found an adversarial example
for $\mathcal{N}_{f}$ in $\mathcal{B}(\textbf{x},d)$.

\subsection{Non-Linear Multiclass Classifiers}

We now extend our technique to non-linear multiclass
classifiers. Analogously to
\secref{subsec:nonlinear-binary-classifiers}, \tool iteratively
approximates the minimum values of functions $g_{ij}$ in
$\mathcal{B}(\textbf{x},d)$.

\paragraph{\textbf{Example.}}
As an example, consider \figref{fig-deep-search-multiclass-nonlinear}
depicting a multiclass classifier $\mathcal{N}_{f} :
\mathbb{R}^{2}\rightarrow C_{3}$, where $f$ is non-linear and
$C_{3}=\{l_{1},l_{2},l_{3}\}$. Assume that $\mathcal{N}_{f}$
classifies input $\textbf{x}_{0}$ in class $l_2$, and thus, so do
non-linear binary classifiers $\mathcal{N}_{g_{21}}$ and
$\mathcal{N}_{g_{23}}$.

Let us also assume that
$g_{21}(\textbf{x}_{0})<g_{23}(\textbf{x}_{0})$. Since $g_{21}$
evaluates to a smaller value than $g_{23}$ for input $\textbf{x}_{0}$,
we consider it more likely to have an adversarial example. In other
words, we first approximate the minimum value of $g_{21}$ because it
is closer to becoming negative for the initial input. \tool treats
$g_{21}$ as linear within $\mathcal{B}(\textbf{x}_{0},d)$ and
generates $\textbf{x}_{1}$. Observe that input $\textbf{x}_{1}$ is not
adversarial. Now, assume that
$g_{21}(\textbf{x}_{1})>g_{23}(\textbf{x}_{1})$. As a result, \tool
tries to find the minimum of function $g_{23}$ in
$\mathcal{B}(\textbf{x}_{0},d)$, also by treating it as linear. It
generates input $\textbf{x}_{2}$, which is an adversarial example for
classifiers $\mathcal{N}_{g_{23}}$ and $\mathcal{N}_{f}$.

\paragraph{\textbf{\tool for non-linear multiclass classifiers.}}
\begin{algorithm}[t!]
	\small \DontPrintSemicolon \SetKwProg{Fn}{Function}{is}{end}
	\SetKwFunction{Max}{PointMaximum} \SetKwFunction{Min}{PointMinimum}
	\SetKwFunction{DeepSearchMulticlass}{DS-Multiclass}
	\KwIn{input $\textbf{x}\in\mathbb{R}^{n}$, initial input
		$\textbf{x}_{\mathit{init}}\in\mathcal{B}(\textbf{x},d)$,\\\hspace{3em}function
		$f:\mathbb{R}^{n}\rightarrow \mathbb{R}^{m}$, distance
		$d\in\mathbb{R}$}
	\KwOut{$\textbf{x}'\in\mathcal{B}(\textbf{x},d)$}
	
	\Fn{\DeepSearchMulticlass{\normalfont{$\textbf{x}, \textbf{x}_{\mathit{init}}, f, d$}} \ }{
		construct intervals $(I_{1},...,I_{n})$ such that $\mathcal{B}(\textbf{x},d)=I_{1}\times...\times I_{n}$\;
		$l_{i}:=\mathcal{N}_{f}(\textbf{x})$\;
		initialize $\textbf{x}_{0}:=\textbf{x}_{\mathit{init}}$ and $k:= 0$\;
		define $\{g_{ij}\ |\ g_{ij}(\textbf{x}) = f_{i}(\textbf{x})-f_{j}(\textbf{x}), 1\leq j\leq m, i\neq j\}$\;
		\Repeat {$\mathcal{N}_{f}(\textbf{x})\neq\mathcal{N}_{f}(\textbf{x}_{k})$, or $k=\texttt{MaxNum}$}{
			$r: =\arg\min_{j} g_{ij}(\textbf{x}_{k})$\;
			$\textbf{x}_{k+1}:= \texttt{ApproxMin}(\textbf{x}_{k}, g_{ir}, (I_{1},...,I_{n}))$\;
			$k:=k+1$\;	}	
		\Return {\normalfont{$\textbf{x}_{k}$}}\;
		
	}
	\caption{\textbf{\tool for multiclass classifiers.}}
	\label{alg:algorithm-multiclass-classifiers}
\end{algorithm}

\algoref{alg:algorithm-multiclass-classifiers} is the general \tool
algorithm for multiclass classifiers. The inputs are the same as for
\algoref{alg:algorithm-binary-classifiers}. For now, assume that
$\textbf{x}$ is equal to $\textbf{x}_{\mathit{init}}$. 
Again, the algorithm executes at most \texttt{MaxNum} iterations, and it terminates as soon as an adversarial example is found. 

Function \texttt{DS-Multiclass} assigns $\textbf{x}_{\mathit{init}}$
to $\textbf{x}_{0}$ and constructs $n$ intervals $I_{1},..., I_{n}$ to
represent $\mathcal{B}(\textbf{x}_{0},d)$. It also computes the class
label $l_i$ of $\textbf{x}$, and defines functions $g_{ij}$
$(1\leq j\leq m, i\neq j)$ (lines 2--5).
The rest of the algorithm uses \texttt{ApproxMin} from
\algoref{alg:algorithm-binary-classifiers} to iteratively approximate
the minimum of one function $g_{ij}$ per iteration, which is
selected on line~7 such that its value for input $\textbf{x}_{k}$ is
smaller in comparison to all other constituent binary classification
functions. Intuitively, $g_{ij}$ corresponds to the binary classifier that is most likely
to have an adversarial example near $\textbf{x}_{k}$. This heuristic allows our algorithm to find an adversarial
example faster than having to generate an input $\textbf{x}_{k+1}$ for
all $m-1$ functions $g_{ij}$ per iteration.

To find an adversarial example in $k$ iterations, we need at most $2n+n(k-1)$ queries for the value of $f$.

\paragraph{\textbf{An alternative objective function.}}
In each iteration of \algoref{alg:algorithm-multiclass-classifiers}, we construct a different objective function $g_{ij}$ and approximate its minimum value. An alternative choice of an objective function is $f_{i}$ itself. In multiclass classification, decreasing the value of $f_{i}$ amounts to decreasing the score value of the $i$th class, which implicitly increases the score
values of other classes. 

We refer to the algorithm derived by substituting lines 7--8 of  \algoref{alg:algorithm-multiclass-classifiers} with the following assignment as \algoref{alg:algorithm-multiclass-classifiers}' :
$$\textbf{x}_{k+1}:= \texttt{ApproxMin}(\textbf{x}_{k}, f_{i}, (I_{1},...,I_{n}))$$

\noindent It uses \texttt{ApproxMin} to iteratively approximate the minimum value of $f_{i}$. We find it very effective in our experiments.
\section{Iterative Refinement}\label{sec:deepsearch-with-refinement}
The closer the adversarial examples are to a correctly classified input,
the more subtle they are. Such adversarial examples are said to have a
low distortion rate. In this section, we extend \tool with an
iterative-refinement approach for finding subtle adversarial examples.
On a high level, given an input $\textbf{x}$ and a distance $d$, for
which we have already found an adversarial example $\textbf{x}'$ in
region $\mathcal{B}(\textbf{x},d)$, \tool iteratively reduces distance
$d$ as long as the smaller region still contains an adversarial
example. If none is found, the distance is not reduced further.

\begin{figure}[t!]
	\centering
	\vspace{-15pt}
	\subfloat[Linear classifier]{{\includegraphics[width=3.8cm]{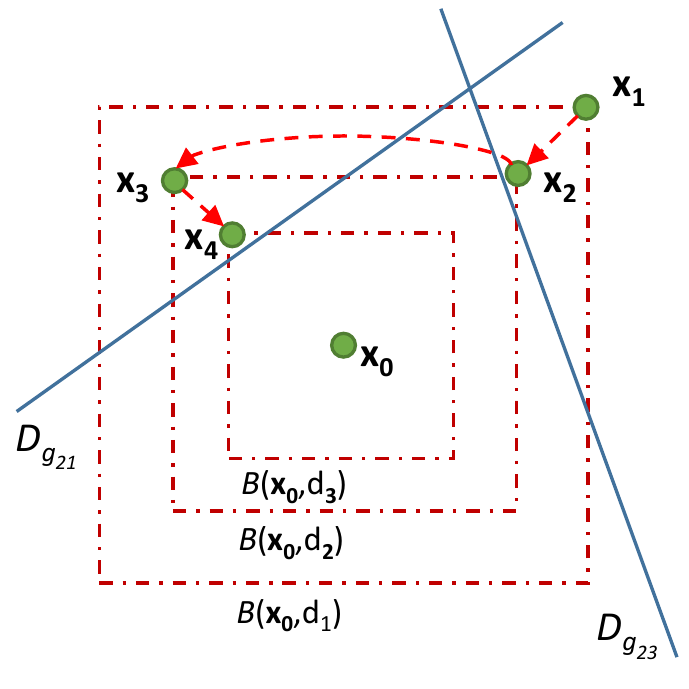}}\label{fig-refinement-linear}}%
	\vspace{-5pt}
	\qquad
	\subfloat[Non-linear classifier]{{\includegraphics[width=3.8cm]{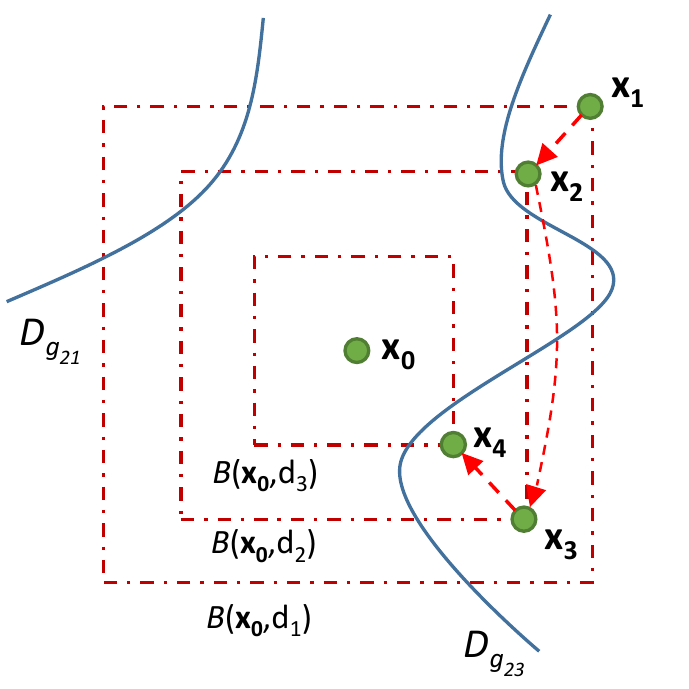}}\label{fig-refinement-nonlinear}}%
	\caption{\tool with iterative refinement.}
	\vspace{-1em}
	\label{fig-refinement}
\end{figure}

Let $\mathcal{I}=I_{1}\times...\times I_{n}$ be an $n$-dimensional
cube, where $I_{i}=[l_{i},u_{i}]$ is a closed interval bounded by
$l_{i},u_{i}\in\mathbb{R}$ with $l_{i}\leq u_{i}$ for $1\leq i\leq n$.
For an input $\textbf{x}$ with an $i$th coordinate
$\textbf{x}(i)\in(-\infty, l_{i}]\cup[u_{i},+\infty)$ for $1\leq i\leq
n$, we define a projection operator $\textsc{Proj}$ that maps
$\textbf{x}$ to a vertex of $\mathcal{I}$ as follows
\begin{displaymath}
	\textsc{Proj}(\mathcal{I},\textbf{x})(i) = \left\{ \begin{array}{ll}
		u_{i}, & \hspace{0.5cm}\textrm{if $\textbf{x}(i)\geq u_{i}$}\\
		l_{i}, & \hspace{0.5cm} \textrm{if $\textbf{x}(i) \leq l_{i}$,}
	\end{array} \right.
\end{displaymath}
where $\textsc{Proj}(\mathcal{I},\textbf{x})(i)$ denotes the $i$th
coordinate of $\textsc{Proj}(\mathcal{I},\textbf{x})$.
As an example, consider \figref{fig-refinement-linear} showing a
linear multiclass classifier. Input $\textbf{x}_{2}$ is a projection
of $\textbf{x}_{1}$.

Using this operator, the minimum and maximum values of
a linear classification function $f$ may also be projected on
$\mathcal{I}$, and we have the following theorem (whose proof is available in \cite{ZhangChowdhury2020}).

\begin{theorem}\label{theorem-projection}
	Let $f(\textbf{x})=\textbf{w}^{T}\textbf{x}+b$ be a linear
	classification function, and $\mathcal{I}_{1}$, $\mathcal{I}_{2}$ two
	$n$-dimensional cubes such that
	$\mathcal{I}_{1}\subseteq\mathcal{I}_{2}$. Assuming that $\textbf{x}$
	is a vertex of
	$\mathcal{I}_{2}$, we have:
	\begin{enumerate}
		\item if $\min$  $f(\mathcal{I}_{2})=f(\textbf{x})$, then $\min$ $f(\mathcal{I}_{1}) = f(\textsc{Proj}(\mathcal{I}_{1},\textbf{x}))$
		\item if $\max$ $f(\mathcal{I}_{2})=f(\textbf{x})$, then $\max$ $f(\mathcal{I}_{1}) = f(\textsc{Proj}(\mathcal{I}_{1},\textbf{x}))$
	\end{enumerate}
\end{theorem}

In \figref{fig-refinement-linear}, assume that input $\textbf{x}_{0}$
is correctly classified in class $l_2$. Then, in region
$\mathcal{B}(\textbf{x}_{0},d_{1})$, function $g_{23}$ obtains its
minimum value for input $\textbf{x}_{1}$. When projecting
$\textbf{x}_{1}$ to vertex $\textbf{x}_{2}$ of
$\mathcal{B}(\textbf{x}_{0},d_{2})$, notice that $g_{23}$ evaluates to
its minimum for input $\textbf{x}_{2}$ in this smaller region.

\paragraph{\textbf{Example.}}
\figref{fig-refinement} shows two multiclass classifiers
$\mathcal{N}_{f} : \mathbb{R}^{2}\rightarrow C_{3}$, where
$C_{3}=\{l_{1},l_{2},l_{3}\}$. In \figref{fig-refinement-linear},
function $f$ is linear, whereas in \figref{fig-refinement-nonlinear},
it is non-linear. For correctly classified input $\textbf{x}_{0}$, we
assume that $\mathcal{N}_{f}(\textbf{x}_{0})=l_{2}$, and thus,
$\mathcal{N}_{g_{21}}(\textbf{x}_{0})=l_{2}$ and
$\mathcal{N}_{g_{23}}(\textbf{x}_{0})=l_{2}$.

In both subfigures, assume that $\textbf{x}_{1}$ is an adversarial
example found by \texttt{DS-Multiclass}
(see \algoref{alg:algorithm-multiclass-classifiers}) in
$\mathcal{B}(\textbf{x}_{0},d_{1})$. Once such an example is found,
our technique with refinement uses bisect search to find the smallest
distance $d'$ such that the projection of $\textbf{x}_{1}$ on
$\mathcal{B}(\textbf{x}_{0},d')$ is still
adversarial. In \figref{fig-refinement}, this distance is $d_2$, and
input $\textbf{x}_{2}$ constitutes the projection of $\textbf{x}_{1}$
on $\mathcal{B}(\textbf{x}_{0},d_{2})$. So, $\textbf{x}_{2}$ is closer
to $\textbf{x}_{0}$, which means that it has a lower distortion rate
than $\textbf{x}_{1}$. In fact, since we are using bisect search to
determine distance $d_2$, $\textbf{x}_{2}$ is the closest adversarial
input to $\textbf{x}_{0}$ that may be generated by projecting
$\textbf{x}_{1}$ on smaller regions.

However, in region $\mathcal{B}(\textbf{x}_{0},d_{2})$, there may be
other vertices that are adversarial and get us even closer to
$\textbf{x}_{0}$ with projection. To find such examples, we
apply \texttt{DS-Multiclass} again, this time starting from input
$\textbf{x}_{2}$ and searching in region
$\mathcal{B}(\textbf{x}_{0},d_{2})$. As a result, we generate
adversarial input $\textbf{x}_{3}$ in the subfigures. Now, by
projecting $\textbf{x}_{3}$ to the smallest possible region around
$\textbf{x}_{0}$, we compute $\textbf{x}_{4}$, which is the
adversarial example with the lowest distortion rate so far.

Assume that applying \texttt{DS-Multiclass} for a third time, starting
from $\textbf{x}_{4}$ and searching in
$\mathcal{B}(\textbf{x}_{0},d_{3})$, does not generate any other
adversarial examples. In this case, our technique returns
$\textbf{x}_{4}$.

\paragraph{\textbf{\tool with iterative refinement.}}
\algoref{alg:algorithm-refinement} describes our technique
with iterative refinement. Each iteration consists of a refinement and
a search step, which we explain next.

The refinement step (lines 3--4) first calculates the $L_{\infty}$
distance $d$ between $\textbf{x}$ and
$\textbf{x}'$. In \figref{fig-refinement}, input $\textbf{x}$ of the
algorithm is $\textbf{x}_0$, and $\textbf{x}'$ is $\textbf{x}_1$. So,
$\textbf{x}'$ is an adversarial input that was generated by our
technique, and consequently, a vertex of $\mathcal{B}(\textbf{x},d)$.
On line 4, we use bisect search to find the minimum distance $d'$ such
that the input derived by projecting $\textbf{x}'$ on
$\mathcal{B}(\textbf{x},d')$ is still
adversarial. In \figref{fig-refinement}, this is distance $d_2$, and
$\textsc{Proj}(\mathcal{B}(\textbf{x},d'),\textbf{x}')$ of the
algorithm corresponds to adversarial input $\textbf{x}_2$ in the
figure.

Note that this refinement is possible because
of \thmref{theorem-projection}, which guarantees that a linear
function $f$ evaluates to its minimum in $\mathcal{B}(\textbf{x},d')$
for the input derived with projection. When $f$ is non-linear, it
might not evaluate to its minimum for adversarial input
$\textsc{Proj}(\mathcal{B}(\textbf{x},d'),\textbf{x}')$. However, it
is still the case that this projected input is closer to $\textbf{x}$,
i.e.,
$||\textbf{x}-\textsc{Proj}(\mathcal{B}(\textbf{x},d'),\textbf{x}')||_{L_{\infty}}\leq||\textbf{x}-\textbf{x}'||_{L_{\infty}}$,
and thus, has a lower distortion rate.

After selecting an input from which to start the search (line 5), the search step (lines 6--9) calls function \texttt{DS-Binary}
(\algoref{alg:algorithm-binary-classifiers}) or \texttt{DS-Multiclass}
(\algoref{alg:algorithm-multiclass-classifiers}), depending on whether
$f$ is a binary classification function. The goal is to search for
another adversarial example (other than
$\textsc{Proj}(\mathcal{B}(\textbf{x},d'),\textbf{x}')$) in region
$\mathcal{B}(\textbf{x},d')$. In \figref{fig-refinement}, an
adversarial input found by this step, when starting the search from
$\textbf{x}_2$, is $\textbf{x}_3$, which is also a vertex of
$\mathcal{B}(\textbf{x}_0,d_2)$. 

On lines 10--13, we essentially check whether the search step was
successful in finding another adversarial input $\textbf{x}''$.
However, \texttt{DS-Binary} and \texttt{DS-Multiclass} might not return
an adversarial example. If they do, like input $\textbf{x}_3$
in \figref{fig-refinement}, the algorithm iterates (line 14). If not,
like when starting the search from input $\textbf{x}_4$ in the figure,
which is a projection of $\textbf{x}_3$ on
$\mathcal{B}(\textbf{x}_0,d_3)$, then we return the projected input
and terminate. 

\begin{algorithm}[t!]
	\small
	\DontPrintSemicolon
	\SetKwProg{Fn}{Function}{is}{end}
	\SetKwFunction{Max}{PointMaximum}
	\SetKwFunction{Min}{PointMinimum}
	\SetKwFunction{DeepSearchRefinement}{DS-Refinement}
	
	\KwIn{input $\textbf{x}\in\mathbb{R}^{n}$, adversarial input $\textbf{x}' \in \mathcal{B}(\textbf{x},d)$ $(d\in\mathbb{R})$,\\\hspace{3em}function
		$f:\mathbb{R}^{n}\rightarrow \mathbb{R}^{m}$}
	
	\KwOut{an adversarial input $\textbf{x}'' \in \mathcal{B}(\textbf{x},d')$ $(d'\leq d)$}
	
	\Fn{\DeepSearchRefinement{\normalfont{$\textbf{x}, \textbf{x}', f$}} \ }{
		\Repeat{\normalfont{$\textbf{x}''$ is not an adversarial example}}{
			$d:=||\textbf{x}-\textbf{x}'||_{L_{\infty}}$\;
			apply bisect search to find the smallest distance $d'\leq d$ such that input $\textsc{Proj}(\mathcal{B}(\textbf{x},d'),\textbf{x}')$ is an adversarial example.\; 
			choose an $\textbf{x}_{\mathit{new}}\in\mathcal{B}(\textbf{x},d')$, from which to start a new search, e.g. $\textbf{x}_{\mathit{new}}=\textsc{Proj}(\mathcal{B}(\textbf{x},d'),\textbf{x}')$.\;
			\eIf{\normalfont{$f$ is binary}}{
				$\textbf{x}'' := \texttt{DS-Binary(}\textbf{x}, \textbf{x}_{\mathit{new}}, f, d' \texttt{)}$\;
			}{
				$\textbf{x}'' := \texttt{DS-Multiclass(}\textbf{x}, \textbf{x}_{\mathit{new}}, f, d' \texttt{)}$\;
			}
			\eIf{\normalfont{$\textbf{x}''$ is an adversarial example}}{
				$\textbf{x}':=\textbf{x}''$\;
			}{
				$\textbf{x}':= \textsc{Proj}(\mathcal{B}(\textbf{x},d'),\textbf{x}')$\;
			}
		}
		
		\Return {\normalfont{$\textbf{x}'$ and $d'$}}\;
		
	}
	\caption{\textbf{\tool with iterative refinement.}}
	\label{alg:algorithm-refinement}
\end{algorithm}
\section{Hierarchical Grouping}\label{sec:deepsearch-hierarchical-grouping}
For an $n$-dimensional input $\textbf{x}$, our technique
makes at least $n$ queries per iteration. For high-dimensional inputs, it could cost a significantly large number of queries to perform even one iteration of our attack. One basic strategy for query reduction is to divide pixels of an input image into different groups and mutate all pixels in a group to the same direction, e.g., all pixels in a group are moved to their upper bounds. Thus, we only need one query for all pixels in the same group. To exploit spatial regularities in images for query efficiency, we adapt
hierarchical grouping \cite{MoonAn2019} to our setting.

\paragraph{\textbf{\tool with hierarchical grouping.}}

\algoref{alg:algorithm-hierarchy} summarizes our technique with hierarchical grouping, which consists of the following three main steps.

\begin{enumerate}
	\item \emph{Initial grouping} (line $8$): For an  $n$-dimensional input image $\textbf{x}$,  we first divide the $n$ dimensions into $\lceil \frac{n}{k^{2}}\rceil$
	sets $G_{1},...,G_{\lceil \frac{n}{k^{2}}\rceil}$, where
	each set $G_{i}$ ($1\leq i\leq \lceil\frac{n}{k^{2}}\rceil$) contains indices corresponding to
	$k\times k$ neighboring pixels in $\textbf{x}$. This amounts to dividing the original image $\textbf{x}$ into $\lceil \frac{n}{k^{2}}\rceil$ square blocks. The definition of $\texttt{Initial-Group(}\{1,...,n\},k\texttt{)}$ is omitted due to space limitations.
	
	\item \emph{Fuzzing} (line $10$): We extend \tool to handle groups of pixels and write $\texttt{DeepSearch(}\textbf{x}, \textbf{x}', f, d, \mathcal{G}\texttt{)}$ to mean such an extension. For each set $G_{i}\in\mathcal{G}$, our technique mutates all coordinates
	that correspond to indices in the set toward the same direction at the
	same time. Hence, \tool only compares two values per set, namely
	$f[u_{i_{1}}/x_{i_{1}},...,u_{i_{l}}/x_{i_{l}}]$ and
	$f[l_{i_{1}}/x_{i_{1}},...,l_{i_{l}}/x_{i_{l}}]$, where
	$i_{1},...,i_{l}\in G_{i}$ and $l=|G_{i}|$.  
	
	\item \emph{Group splitting} (line $11$--$13$): If the current partition of the image is still too coarse for \tool to find adversarial examples, we perform \tool in finer granularity. We further divide each set $G_{i}$ into $m\times m$ subsets $G_{i,1},...,G_{i,m\times m}$, where each set $G_{i,j}$ ($1\leq j\leq m\times m$) contains indices corresponding to $k/m\times k/m$ neighboring pixels in $\textbf{x}$. After splitting all sets, the total number of sets is multiplied by $m\times m$. This results in a more fine-grained partition of input $\textbf{x}$. We then go back to step $(2)$. 
	
\end{enumerate}

In the query-limited setting, we use single-step \tool on line $10$, i.e., we fix \texttt{MaxNum} to 1 in \algoref{alg:algorithm-multiclass-classifiers} and \algoref{alg:algorithm-multiclass-classifiers}', and choose $\textbf{x}_{\mathit{init}}$ to be a vertex of $\mathcal{B}(\textbf{x},d)$ to avoid unnecessary queries. Hence, when there are $\lceil \frac{n}{k^{2}}\rceil$ sets in $\mathcal{G}$, the total number of queries per iteration in \algoref{alg:algorithm-hierarchy} reduces to $\lceil \frac{n}{k^{2}}\rceil$.

\begin{algorithm}[t!]
	\small
	\DontPrintSemicolon
	\SetKwProg{Fn}{Function}{is}{end}
	\SetKwFunction{DivideGroup}{Divide-Group}
	\SetKwFunction{DeepSearchGrouping}{DS-Hierarchy}
	
	\KwIn{input $\textbf{x}\in\mathbb{R}^{n}$, initial input
		$\textbf{x}_{\mathit{init}}\in\mathcal{B}(\textbf{x},d)$, initial group size $k$, parameter $m$ for group splitting, function
		$f:\mathbb{R}^{n}\rightarrow \mathbb{R}^{m}$, distance
		$d\in\mathbb{R}$, query budget $L$}
	
	\KwOut{$\textbf{x}'\in\mathcal{B}(\textbf{x},d)$}
	
	\Fn{\DivideGroup{\normalfont{$\mathcal{G}, m$}} \ }{
		\ForEach{$G_{i}\in\mathcal{G}$}{
			divide $G_{i}$ into $m\times m$ subset $\{G_{i,1},...,G_{i,m\times m}\}$\;
			$\mathcal{G}':= \mathcal{G}\cup \{G_{i,1},...,G_{i,m\times m}\}$
		}
		
		\Return{\normalfont{$\mathcal{G}'$}}
	}
	\;
	\Fn{\DeepSearchGrouping{\normalfont{$\textbf{x}, \textbf{x}_{\mathit{init}}, f, k, m$}} \ }{
		$\mathcal{G}:=\texttt{Initial-Group(}\{1,...,n\},k\texttt{)}$ and $\textbf{x}':= \textbf{x}_{\mathit{init}}$\;
		\Repeat{$\mathcal{N}_{f}(\textbf{x})\neq\mathcal{N}_{f}(\textbf{x}')$, or \normalfont{reached query budget $L$}}{

			$\textbf{x}':=\texttt{DeepSearch(}\textbf{x}, \textbf{x}', f, d, \mathcal{G}\texttt{)}$\;

			\If{\normalfont{$1<k/m$}}{
				$\mathcal{G}:=\texttt{Divide-Group(}\mathcal{G},m\texttt{)}$\;
				$k:= k/m$
			}
			
		}
		
		\Return {\normalfont{$\textbf{x}'$}}\;
		
	}
	\caption{\textbf{\tool with hierarchical grouping.}}
	\label{alg:algorithm-hierarchy}
\end{algorithm}
\section{Experimental Evaluation}\label{sec:Experiments}
We evaluate \tool by using it to test the robustness
of deep neural networks trained for popular datasets. We also compare
its effectiveness with state-of-the-art blackbox
attacks. Our experiments are designed around the following research
questions:

\begin{description}
	\item[RQ1:] Is \tool effective in finding adversarial examples?
	\item[RQ2:] Is \tool effective in finding adversarial examples with low distortion rates?
	\item[RQ3:] Is \tool a query-efficient blackbox attack?
	\item[RQ4:] Is the hierarchical grouping of \tool effective in improving query efficiency?
\end{description}

We make our implementation open source\footnote{\url{https://github.com/Practical-Formal-Methods/DeepSearch}}. Our experimental data, including detected adversarial examples, are
also available via the provided link.

\subsection{Evaluation Setup}
\label{subsec:Setup}

\paragraph{\textbf{Datasets and network models.}} 

We evaluate our approach on deep neural networks trained for three well known
datasets, namely SVHN \cite{NetzerWang11} (cropped digits), CIFAR-$10$ \cite{Krizhevsky2008}, and ImageNet \cite{RussakovskyDeng15}. 
For each dataset, we randomly selected $1000$ correctly classified images from the
test set on which to perform blackbox attacks.

For SVHN and CIFAR-$10$, we attack two wide ResNet w$32$-$10$ \cite{ZagoruykoKomodakis16} networks, where one is naturally trained (without defense) and the other is adversarially trained with a state-of-the-art defense \cite{MadryMakelov2018}. For SVHN, the undefended network we trained has $95.96\%$ test accuracy, and the adversarially trained network has $93.70\%$ test accuracy. During adversarial training, we used the PGD attack \cite{MadryMakelov2018} (that can perturb each pixel by at most
$8$ on the $0$--$255$ pixel scale) to generate  adversarial examples. For CIFAR-$10$, we trained an undefended network with $95.07\%$ test accuracy. The defended network we attack is the pretrained network provided in Madry's challenge\footnote{\url{https://github.com/MadryLab/cifar10_challenge}}. For ImageNet, we attack a pretrained Inception v3 network \cite{SzegedyVanhoucke2016}, which is undefended. 

Defenses for ImageNet networks are also important, and we would have attacked defended ImageNet networks
if they were publicly available. In this work, we did not attack such networks (using the defense in \cite{MadryMakelov2018}) for the following
reasons. First, there are no publicly available ImageNet networks that use the defense in \cite{MadryMakelov2018}. Second, none of the state-of-the-art
attacks that we used for comparison in our paper (i.e., \cite{IlyasEngstrom2018,IlyasEngstrom19,GuoGardner19,MoonAn2019}) have been
evaluated on defended networks for this dataset. Therefore, we did not compare \tool with these attacks on such networks.
Third, due to the extremely high computational cost, implementing the defense in \cite{MadryMakelov2018} for an ImageNet network is impractical.

\paragraph{\textbf{Existing approaches.}}
We compare \tool with four state-of-the-art blackbox attacks for generating
adversarial examples:

\begin{itemize}
	
	\item[$\bullet$] The NES attack \cite{IlyasEngstrom2018}, optimized for the $L_{\infty}$ distance metric, is developed for various settings, including a query-limited setting. It uses natural evolution strategies (NES) \cite{SalimansHo2017} for gradient estimation and performs projected gradient-descent (PGD) \cite{MadryMakelov2018} style adversarial attacks using estimated gradients. We compare with the NES attack developed for a query-limited setting, i.e., QL-NES.
	
	\item[$\bullet$] The Bandits attack \cite{IlyasEngstrom19} extended gradient-estimation-based blackbox attacks by integrating gradient priors, e.g., time-dependent and data-dependent priors, through a bandit optimization framework. The Bandits attack can perform both $L_{2}$ and $L_{\infty}$ attacks.
	
	\item[$\bullet$] The Simple Blackbox Attack \cite{GuoGardner19} is optimized for the $L_{2}$ distance metric. Starting from an input image, it finds adversarial examples by repeatedly adding or subtracting a random vector sampled from a set of predefined orthogonal candidate vectors.  We compare with their SimBA algorithm, which can also be easily constrained using $L_{\infty}$ distance. 
	
	\item[$\bullet$] The Parsimonious blackbox attack \cite{MoonAn2019}, optimized for the $L_{\infty}$ distance metric, encodes the problem of finding adversarial perturbations as finding solutions to linear programs. For an input $\textbf{x}$ and distance $d$, it searches among the vertices of $\mathcal{B}(\textbf{x},d)$ and finds adversarial examples by using efficient algorithms in combinatorial optimization.
	
\end{itemize}

\paragraph{\textbf{\tool implementation.}}
We briefly introduce some implementation details of \tool.
In our implementation, the classification function $f$ of multiclass classifiers maps input images to the logarithm of class probabilities predicted by neural networks. In this setting, the objective function in \algoref{alg:algorithm-multiclass-classifiers} (resp. \algoref{alg:algorithm-multiclass-classifiers}') corresponds to logit loss \cite{CarliniWagner2017-Robustness} (resp. cross-entropy loss \cite{DeepLearning}).

To reduce the number of queries, for input $\textbf{x}$ and distance $d$, we choose $\textbf{x}_{\mathit{init}}\in\mathcal{B}(\textbf{x},d)$ such that it is derived from $\textbf{x}$ by setting the values of all its pixels to the lower bounds in  $\mathcal{B}(\textbf{x},d)$. In the refinement step, when calculating new adversarial examples within smaller distances, we set $\textbf{x}_{\mathit{new}}$ to $\textbf{x}_{\mathit{init}}$ for convenience.

In our experiments, we used \algoref{alg:algorithm-multiclass-classifiers} to attack the undefended networks. We used \algoref{alg:algorithm-multiclass-classifiers}' to attack the defended networks since they are more robust against cross-entropy model attacks. To attack the SVHN and CIFAR-$10$ networks, we set the initial grouping size to $4\times4$. For ImageNet, we set the initial grouping size to $32\times32$ due to their large image size. For group splitting, we set $m=2$ so that we always divide a group into $2\times2$ subgroups.

In the hierarchical-grouping setting, we mutate groups of pixels in random orders. To avoid overshooting query budgets, we mutate groups in batches, and the batch size is $64$ in all our experiments.

\paragraph{\textbf{Parameter settings.}}
For all datasets, we set $L_{\infty}$ distance $d=8$ on the $0$--$255$ pixel scale to perform attacks. For both SVHN and CIFAR-$10$ networks, we set the query budget to $20,000$. For the ImageNet network, we set the query budget to $10,000$ as done in related work \cite{IlyasEngstrom19,MoonAn2019} .

For the QL-NES attack, we set $\sigma=0.001$, size of NES population $n=100$, learning rate $\eta=0.001$, and momentum $\beta=0.9$ for SVHN and CIFAR-$10$. We set $\sigma=0.01$, size of NES population $n=100$, learning rate $\eta=0.0001$, and momentum $\beta=0.9$ for ImageNet.

For the Bandits attack, we set OCO learning rate $\eta=0.001$, image learning rate $h=0.0001$, bandits exploration $\delta=0.1$, finite difference probe $\eta=0.1$, and tile size to $16$ for SVHN and CIFAR-$10$. We set OCO learning rate $\eta=1$, image learning rate $h=0.0001$, bandits exploration $\delta=1$, finite difference probe $\eta=0.1$, and tile size to $64$ for ImageNet.

For the Parsimonious attack, we use the parameters mentioned in their paper for CIFAR-$10$ and ImageNet networks. For SVHN, we use the same parameters as for CIFAR-$10$. Moreover, their implementation offers both cross-entropy and logit loss to construct attacks. We tried both loss functions in our experiments and select the one with better performance for comparison.

\subsection{Metrics}
\label{subsec:Metrics}

In our evaluation, we use the following
metrics.

\paragraph{\textbf{Success rate.}}
The success rate measures the percentage of input images for
which adversarial examples are found. The higher this rate, the more
effective a given technique in finding adversarial examples. Assume
we write $\mathit{findAdv}(\textbf{x})$ to denote whether an adversarial
example is found for input $\textbf{x}$. If so, we have that $\mathit{findAdv}(\textbf{x})=1$; otherwise, we have $\mathit{findAdv}(\textbf{x})=0$. For a set of images
$\textbf{X}=\{\textbf{x}_{1},...,\textbf{x}_{k}\}$, the
success rate of a given technique is:

$$\mathit{ASR}(\textbf{X})=\frac{1}{k}\sum_{i=1}^{k}\mathit{findAdv}(\textbf{x}_i)$$

\paragraph{\textbf{Average distortion rate.}}
Let sets $\textbf{X}=\{\textbf{x}_{1},...,\textbf{x}_{k}\}$ and
$\textbf{X}_{\mathit{adv}}=\{\textbf{x}_{1}',...,\textbf{x}_{k}'\}$ be
input images and adversarial examples, respectively. The average
distortion rate between $\textbf{X}$ and $\textbf{X}_{\mathit{adv}}$
with respect to the $L_{\infty}$ distance is:

$$\mathit{AvgDR}_{L_{\infty}}(\textbf{X},\textbf{X}_{\mathit{adv}})=\frac{1}{k}\sum_{i=1}^{k}\dfrac{||\textbf{x}_i-\textbf{x}'_i||_{L_{\infty}}}{||\textbf{x}_i||_{L_{\infty}}}$$

As shown here, the lower this rate, the more subtle the adversarial
examples. For approaches that achieve similar misclassification rates,
we use this metric to determine which approach finds more subtle
perturbations. The average distortion rate with respect to $L_{2}$ can be derived by substituting $L_{\infty}$ with $L_{2}$ in the above definition. In our experimental results, we also include the average $L_{2}$ distortion for reference.

\paragraph{\textbf{Average queries.}} For blackbox attacks, we use the number of queries required to find adversarial examples to measure their efficiency. An approach that requires more queries to perform attacks is more costly.
For each approach, we calculate the average number of queries required to perform successful attacks. We also list their mean number of queries for reference. We point out that queries made by the refinement step of \tool are not counted when calculating average queries because refinement starts after adversarial examples are already found.

\begin{figure*}[t]
	\centering
	\subfloat[SVHN (defended)]{{\includegraphics[width=5.5cm, trim=0cm 2cm 0cm 1cm]{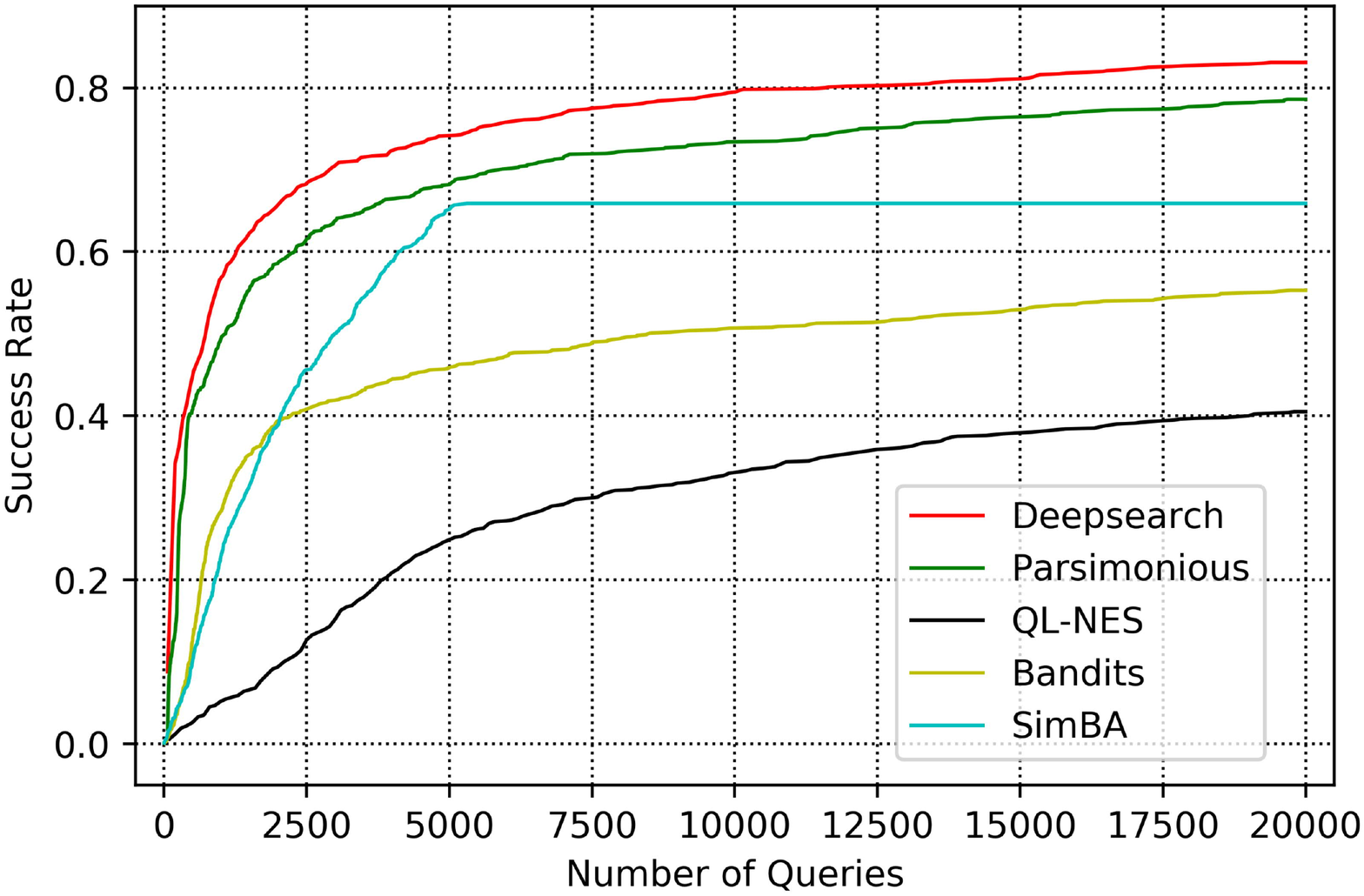}}\label{fig-attack-success-rate-svhn-defended}}
	\qquad
	\subfloat[CIFAR-$10$ (defended)]{{\includegraphics[width=5.5cm, trim=0cm 2cm 0cm 1cm]{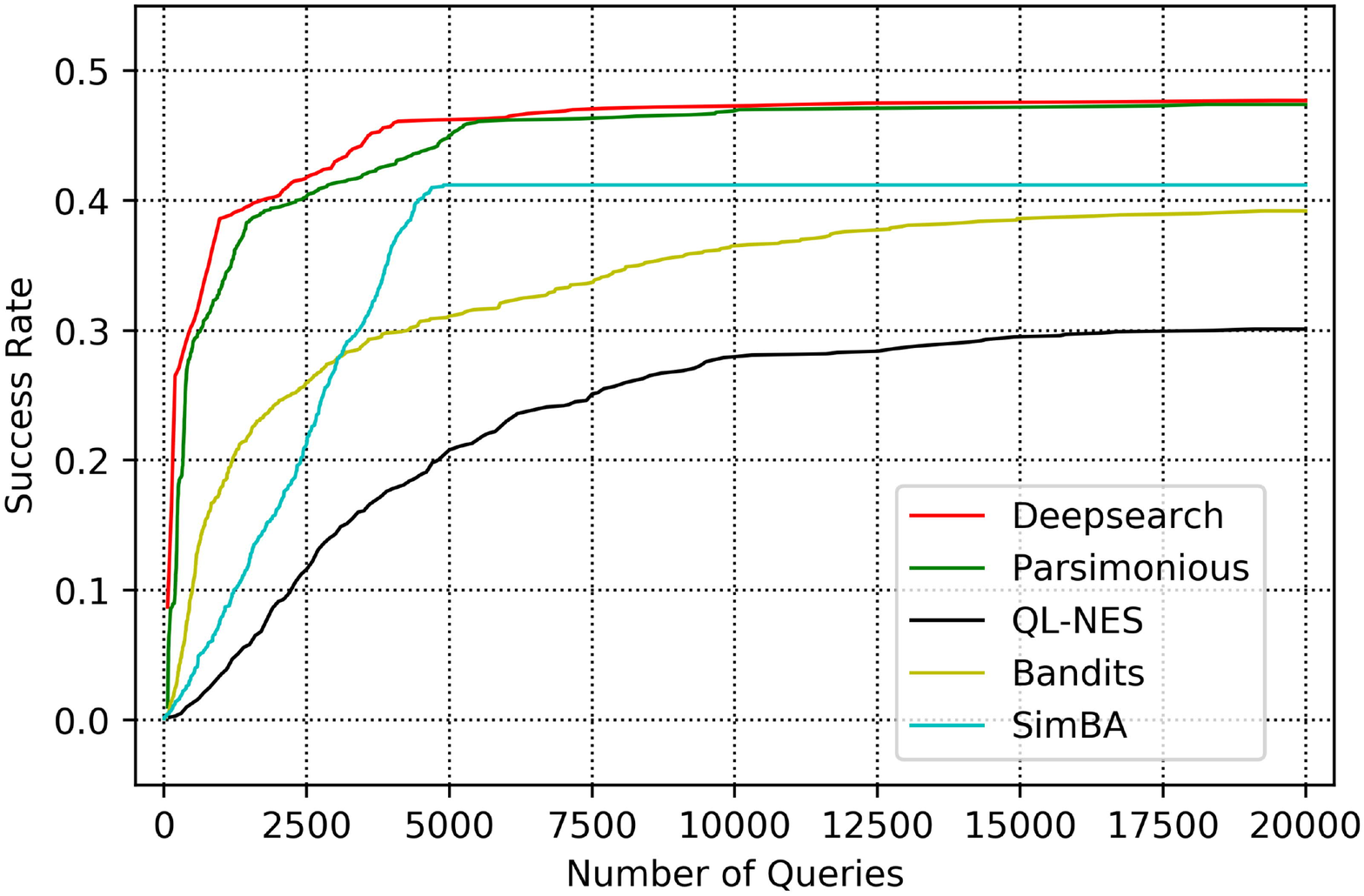}}\label{fig-attack-success-rate-cifar-defended}}
	\qquad
	\subfloat[ImageNet (undefended)]{{\includegraphics[width=5.5cm, trim=0cm 2cm 0cm 1cm]{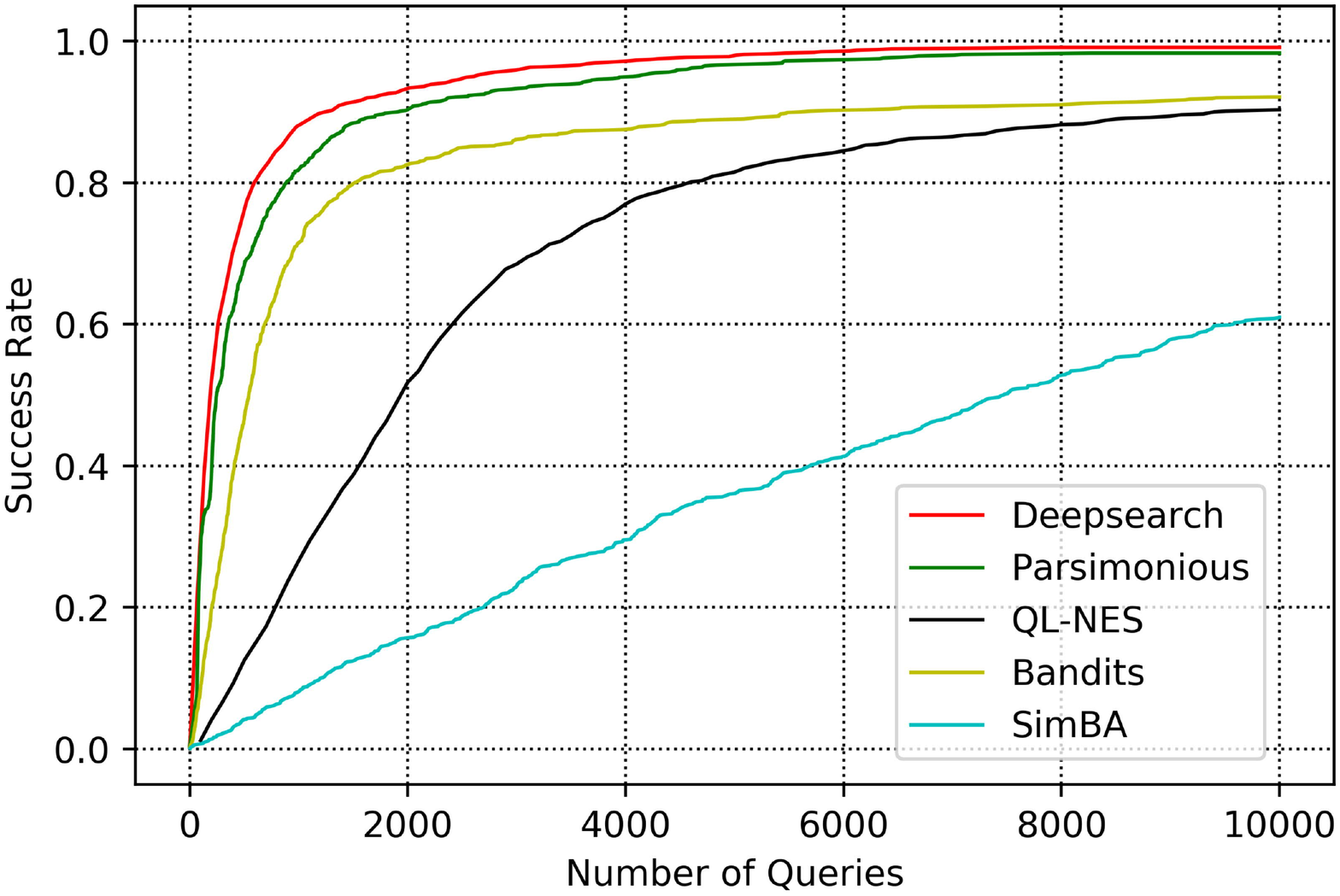}}\label{fig-attack-success-rate-imagenet-undefended}}
	\vspace{-1em}
	\caption{Results on success rate w.r.t number of queries.}
	\label{fig-success-rate-all-dataset}
\end{figure*}

\subsection{Experimental Results}
\label{subsec:Results}

\begin{table}
	\caption{Results on SVHN networks.}
	\vspace{-1em}
	\scalebox{0.95}{
		\begin{tabular}{>{\centering}m{1.4cm} >{\centering}m{1cm} >{\centering}m{1.0cm} >{\centering}m{1cm} >{\centering}m{1.2cm} >{\centering\arraybackslash}m{1cm}}
			\toprule[1pt]
			\textbf{Attack} & \textbf{Success rate} & \textbf{Avg. $L_{\infty}$}& \textbf{Avg. \\ $L_{2}$} & \textbf{Avg. queries}& \textbf{Med. queries}\\
			\hline
			\multicolumn{6}{c}{Undefended network}\\
			\hline	    
			QL-NES & 62.4\% & 2.58\% &1.80\% & 2157 & 1700 \\
			Bandits & 99.2\% & 3.43\% & 2.69\%& 762 & 573 \\
			SimBA & 84.7\% & 4.65\% & 3.47\%& 1675 & 1430 \\
			Parsimonious & 100\% & 4.59\% &7.63\%& 337 & 231 \\
			\textbf{\tool} & \textbf{100\%} & \textbf{1.89\%} &\textbf{3.17\%}& \textbf{229} & \textbf{196} \\	  
			\hline 
			\multicolumn{6}{c}{Defended network}\\
			\hline	    
			QL-NES & 40.5\% & 4.10\% &4.19\% & 5574 & 3900 \\
			Bandits & 55.3\% & 4.38\% &4.74\%& 2819 & 944 \\
			SimBA & 65.9\% & 4.96\% &3.95\%& 2687 & 2633\\
			Parsimonious & 78.9\% & 4.86\% &8.08\%& 2174 & 423.5\\
			\textbf{\tool} & \textbf{83.1\%} & \textbf{3.35\%} &\textbf{5.58\%}& \textbf{1808} & \textbf{458} \\
			\bottomrule[1pt]
	\end{tabular}}
	\vspace{-1em}
	\label{table-svhn-networks}
\end{table}

\begin{table}
	\caption{Results on CIFAR-$10$ networks.}
	\vspace{-1em}
	\begin{tabular}{>{\centering}m{1.4cm} >{\centering}m{1cm} >{\centering}m{1.0cm} >{\centering}m{1.0cm}>{\centering}m{1cm} >{\centering\arraybackslash}m{1cm}}
		\toprule[1pt]
		\textbf{Attack} & \textbf{Success rate} & \textbf{Avg. $L_{\infty}$} & \textbf{Avg. \\ $L_{2}$} & \textbf{Avg. queries} & \textbf{Med. queries}\\
		\hline
		\multicolumn{6}{c}{Undefended network}\\
		\hline	    
		QL-NES & 52.8\% & 1.24\% &0.99\%& 1360 & 1100 \\
		Bandits & 92.6\% & 2.66\% &2.34\%& 838 & 616 \\
		SimBA & 71.6\% & 3.36\% &2.19\%& 1311 & 1150 \\
		Parsimonious & 100\% & 3.36\% &6.36\%& 339& 238.5\\
		\textbf{\tool} & \textbf{100\%} & \textbf{1.64\%} &\textbf{3.08\%}& \textbf{247} & \textbf{196} \\	    
		\hline 
		\multicolumn{6}{c}{Defended network}\\
		\hline	    
		QL-NES & 30.1\% & 2.71\% &3.09\%& 4408 & 3200 \\
		Bandits & 39.2\% & 2.95\% &4.39\%& 2952 & 1176 \\
		SimBA & 41.2\% & 3.46\% &4.50\%& 2425 & 2424 \\
		Parsimonious & 47.4\% & 3.45\% &6.61\%& 1228 & 366 \\
		\textbf{\tool} & \textbf{47.7\%} & \textbf{2.48\%} &\textbf{4.70\%}& \textbf{963} & \textbf{196}\\
		\bottomrule[1pt]
	\end{tabular}
	\vspace{-0.5em}
	\label{table-cifar-networks}
\end{table}

\begin{table}
	\caption{Results on ImageNet undefended network.}
	\vspace{-1em}
	\begin{tabular}{>{\centering}m{1.4cm} >{\centering}m{1cm} >{\centering}m{1.0cm}>{\centering}m{1.0cm} >{\centering}m{1cm} >{\centering\arraybackslash}m{1cm}}
		\toprule[1pt]
		\textbf{Attack} & \textbf{Success rate} & \textbf{Avg. $L_{\infty}$} &\textbf{Avg. \\ $L_{2}$} & \textbf{Avg. queries} & \textbf{Med. queries}\\
		\hline
		QL-NES & 90.3\% & 1.83\% &1.75\%& 2300 & 1800 \\
		Bandits & 92.1\% & 2.15\% &2.61\%& 930 & 496 \\
		SimBA & 61\% & 3.15\% &0.67\%& 4379 & 4103\\
		Parsimonious & 98.3\% & 3.16\% &6.35\%& 660 & 241 \\
		\textbf{\tool} & \textbf{99.3\%} & \textbf{1.50\%} &\textbf{3.05\%}&\textbf{561} & \textbf{196} \\
		\bottomrule[1pt]
	\end{tabular}
	\vspace{-0.5em}
	\label{table-imagenet-undefended}
\end{table}

\tool outperforms all other blackbox attacks in success rate, average queries, and average distortion rate. Experimental results are shown in Fig.~\ref{fig-success-rate-all-dataset} and Tabs.~\ref{table-svhn-networks}--\ref{table-query-reduction-imagenet}.  

\paragraph{\textbf{Results on success rate (RQ1).}}
\tool is very effective in finding adversarial examples for both undefended and defended networks. Compared to other blackbox attacks, \tool has the highest attack success rate.

For undefended networks, the success rate of \tool is close to $100\%$ for all three datasets. For the SVHN defended network, \tool has a success rate of $83.1\%$, which is $4.2\%$ higher than that of the Parsimonious attack (the second best attack in success rate). For the CIFAR-$10$ defended network, \tool has a success rate of $47.7\%$. 

\paragraph{\textbf{Results on average distortion rate (RQ2).}}
\tool has found adversarial examples with the lowest average $L_{\infty}$ distortion rate for networks of all datasets. For the CIFAR-$10$ undefended network, QL-NES has a success rate of only $52.8\%$. To have a more fair comparison, for those $52.8\%$ images that are successfully attacked by both \tool and QL-NES, we further calculate the average distortion rate of the adversarial examples found by \tool. We find it to be $1.2\%$, which is lower than that of QL-NES.  Although \tool is an $L_{\infty}$ attack, adversarial examples found by \tool also have low average $L_{2}$ distortion rate.

\paragraph{\textbf{Results on query efficiency (RQ3).}}
\tool outperforms all other attacks in query efficiency. Compared to the Parsimonious attack (the second best attack in query efficiency), \tool reduces the average queries by $15$--$32\%$ across all datasets.

\paragraph{\textbf{Results on query reduction (RQ4).}}
\begin{table}
	\caption{Query reduction for SVHN and CIFAR-$10$. For each dataset, success rate (resp. average queries) is shown in the first (resp. second) row.}
	\vspace{-1em}        
	\begin{tabular}{>{\centering}m{1.3cm} >{\centering}m{1cm} >{\centering}m{1cm} >{\centering}m{1cm} >{\centering}m{1cm} >{\centering\arraybackslash}m{1cm}}
		\toprule[1pt]
		\textbf{Dataset} & \textbf{1} & \textbf{2$\times$2} & \textbf{4$\times$4} & \textbf{8$\times$8} & \textbf{16$\times$16}\\
		\hline
		\multicolumn{6}{c}{Undefended network}\\
		\hline	    
		\multirow{2}{*}{SVHN} 
		& 100\%  & 100\% & 100\% & 100\% & 100\% \\
		& 742 & 300  & 229 & 238 & 242 \\
		\cmidrule{2-6}
		\multirow{2}{*}{CIFAR-10}    
		& 100\% & 100\% & 100\% & 100\% & 100\% \\
		& 462 & 301 & 247 & 255 & 259 \\
		\hline
		\multicolumn{6}{c}{Defended network}\\  
		\hline
		\multirow{2}{*}{SVHN} 
		& 81.3\%  & 82.4\% & 83.1\% & 83.6\% & 83.9\% \\
		& 3143  & 2292 & 1808 & 1565 & 1591 \\
		\cmidrule{2-6}
		\multirow{2}{*}{CIFAR-10}    
		& 47.7\% & 47.4\% & 47.7\% & 47.6\% & 47.6\% \\
		& 2292 & 1156 & 963 & 935 & 946 \\	
		\bottomrule[1pt]
	\end{tabular}
	\vspace{-0.5em}        
	\label{table-query-reduction-svhn-cifar}
\end{table}

\begin{table}
	\caption{Query reduction for ImageNet. Success rate (resp. average queries) is shown in the first (resp. second) row.}
	\vspace{-1em}    
	\begin{tabular}{>{\centering}m{1.3cm} >{\centering}m{1cm} >{\centering}m{1cm} >{\centering}m{1cm} >{\centering}m{1cm} >{\centering\arraybackslash}m{1cm}}
		\toprule[1pt]
		\textbf{Dataset} & \textbf{8$\times$8} & \textbf{16$\times$16} & \textbf{32$\times$32} & \textbf{64$\times$64} & \textbf{128$\times$128} \\
		\hline			    		
		\multirow{2}{*}{ImageNet}
		& 99.1\% & 99.1\% & 99.1\% & 99.4\%  & 99.4\% \\
		& 765 & 580 & 533 & 554 & 580 \\		
		\bottomrule[1pt]
	\end{tabular}	
	\vspace{-0.5em}        
	\label{table-query-reduction-imagenet}
\end{table}

We demonstrate the effectiveness of hierarchical grouping in \tool for query reduction. We use \tool to attack networks with different initial group sizes and show their corresponding success rates and average queries in Tabs.~\ref{table-query-reduction-svhn-cifar} and \ref{table-query-reduction-imagenet}.

We first notice that the initial group size can only slightly affect attack success rate. For the SVHN defended network, the success rate increases from $81.3\%$ to $83.9\%$ as initial group size increases from $1$ to $16\times16$. However, the changes in success rate are negligible for all other networks.

On the other hand, we observe that hierarchical grouping improves query efficiency dramatically. We take the average queries of group size $1$ and $4\times4$ as an example for SVHN and CIFAR-$10$ networks. For the SVHN undefended network, we see a $69.1\%$ decrease of average queries from $742$ to $229$.  For the SVHN defended network, average queries are reduced by $42.5\%$ from $3143$ to $1808$. For the CIFAR-$10$ undefended network, the average queries are decreased by $46.5\%$ from $462$ to $247$. For the CIFAR-$10$ defended network, average queries are decreased by $58\%$ from $2292$ to $963$, and for the ImageNet network, from group size $8\times8$ to $32\times32$, we decreased the average queries by $30.3\%$ from $765$ to $533$.

\subsection{Threats to Validity}
\label{subsec:Threats}

We have identified the following three threats to the validity of our
experiments.

\paragraph{\textbf{Datasets and network models.}}
Our experimental results may not generalize to other datasets or
network models. However, we used three of the most popular datasets for
image classification, SVHN, CIFAR-10, and ImageNet. Moreover, our network models
have very high test accuracy and the defense we use based on adversarial training is state of the art.

\paragraph{\textbf{Existing approaches.}}
The second threat is related to the choice of existing approaches with
which we compare.
\tool uses iterative linearization of non-linear neural networks and is tailored to the $L_{\infty}$ distance metric. We, thus, compare with
approaches that can also perform $L_{\infty}$ attacks. To our knowledge, the blackbox attacks with which we compared are all state-of-the-art $L_{\infty}$ attacks.

\paragraph{\textbf{Fairness of comparison.}}
The selection of parameters for each approach could affect the
fairness of our comparison. We tried various parameters for each attack and choose the ones yielding best performance. 
\section{Related Work}\label{sec:RelatedWork}
\paragraph{\textbf{Adversarial robustness.}}

Szegedy et al. \cite{SzegedyZaremba2014} first discovered adversarial examples in neural networks and used box-constrained L-BFGS to find them. Since then, multiple whitebox adversarial attacks have been proposed: FGSM \cite{GoodfellowShlens2015}, BIM \cite{KurakinGoodfellow2017}, DeepFool \cite{MoosaviDezfooliFawzi2016}, JSMA \cite{PapernotMcDaniel2016-Limitations}, PGD \cite{MadryMakelov2018}, and  C\&W \cite{CarliniWagner2017-Robustness}. Goodfellow et al. \cite{GoodfellowShlens2015} first argued that the primary cause of adversarial examples is the linear nature of neural networks, and they proposed FGSM that allows fast generation of adversarial examples. BIM improved FGSM by extending it with iterative procedures. DeepFool \cite{MoosaviDezfooliFawzi2016} is another method that performs adversarial attacks through iterative linearization of neural networks.

Blackbox adversarial attacks are more difficult than whitebox ones, and many blackbox attacks require a large number of queries. Papernot et al. \cite{PapernotMcDaniel2016-Transferability,PapernotMcDaniel2017} explored blackbox attacks based on the phenomenon of transferability \cite{SzegedyZaremba2014,PapernotMcDaniel2016-Transferability}. Chen et al. \cite{ChenZhang2017} and  Bhagoji et al. \cite{BhagojiHe2018} proposed blackbox attacks based on gradient estimation \cite{StochasticSearch,Calculus}.  Uesato et al. \cite{UesatoODonoghue2018} used SPSA \cite{Spall1992}.  Ilyas et al. \cite{IlyasEngstrom2018,IlyasEngstrom19} used NES \cite{SalimansHo2017} and proposed the Bandits attack. Narodytska et al. \cite{NarodytskaKasiviswanathan2017} performed a local-search-based attack. The boundary attack \cite{BrendelRauber2018} only requires access to the final decision of neural networks. Guo et al. \cite{GuoGardner19} further considered perturbations in low frequency space. Moon et al. \cite{MoonAn2019} leveraged algorithms in combinatorial optimization.

Although research on developing adversarial attacks is moving fast, research on defending neural networks against adversarial attacks is relatively slow \cite{CarliniWagner2017-Robustness,Cornelius2019,LuChen2018,SharmaChen2018,AthalyeCarlini2018-CVPR,EngstromIlyas2018,AthalyeCarlini2018-Gradients,CarliniWagner2017-MagNet,HeWei2017,CarliniWagner2017-Bypassing,CarliniWagner2016}. Many defense techniques are shown to be ineffective soon after they have been developed. We refer to the work of Carlini et al. \cite{CarliniAthalye2019} for a more detailed discussion on evaluating adversarial robustness.

\paragraph{\textbf{Testing deep neural networks.}} Recently, significant progress has been made on testing neural networks. Several useful test coverage criteria have been proposed to guide test case generation: DeepXplore \cite{PeiCao2017} proposed neuron coverage and the  first whitebox testing framework for neural networks; DeepGauge\cite{MaJuefeiXu2018} proposed a set of finer-grained test coverage criteria; DeepCT \cite{MaZhang2018} further proposed combinatorial test coverage for neural networks; Sun et al. \cite{SunHuang2018} proposed coverage criteria inspired by MC/DC; Kim et al. \cite{KimFeldt2019} proposed surprise adequacy for deep learning systems. Sekhon et al.~\cite{SekhonFleming2019} and Li et al. \cite{LiMa2019} pointed out the limitation of existing structural coverage criteria for neural networks. Li et al. \cite{SekhonFleming2019} also discussed improvements for better coverage criteria.

Moreover, Sun et al. \cite{SunWu2018} proposed the first concolic testing \cite{SenMarinov2005,GodefroidKlarlund2005} approach for neural networks. DeepCheck \cite{GopinathWang2018}  tests neural networks based on symbolic execution \cite{Clarke1976,King1976}. TensorFuzz \cite{OdenaOlsson2019} proposed the first framework of coverage-guided fuzzing for neural networks. DeepHunter \cite{XieMa2019} considered various mutation strategies for their fuzzing framework. Wicker et al. \cite{WickerHuang2018} extracted features from images and computed adversarial examples using a two-player turn-based stochastic game. DLFuzz \cite{GuoJiang2018} proposed the first differential fuzzing framework for deep learning systems. DeepTest \cite{TianPei2018} and DeepRoad \cite{ZhangZhang2018} proposed testing tools for autonomous driving systems based on deep neural networks. For more on testing neural networks, we refer to the work of Zhang et al. \cite{ZhangHarman2019} that surveys testing of machine-learning systems.

\paragraph{\textbf{Formal verification of deep neural networks.}} Verification of neural networks is more challenging than testing. Early work \cite{PulinaTacchella2010} used abstract interpretation~\cite{CousotCousot1977} to verify small-sized neural networks. Recent work \cite{GopinathKatz2018,KatzHuang2019,SinghGehr2019} used SMT \cite{BarrettTinelli2018} techniques and considered new abstract domains. 

Liu et al. \cite{LiuArnon2019} classified recent work in the area into five categories: Reachability-analysis based approaches include MaxSens \cite{XiangTran2018}, ExactReach \cite{XiangTran2017}, and AI$^{2}$\cite{GehrMirman2018}; NSVerify \cite{LomuscioMaganti2017}, MIPVerify \cite{TjengXiao2019} and ILP\cite{BastaniIoannou2016} are based on primal optimization; Duality \cite{DvijothamStanforth2018}, ConvDual \cite{WongKolter2018} and Certify \cite{RaghunathanSteinhardt2018} use dual optimization; Fast-Lin and Fast-Lip \cite{WengZhang2018}, ReluVal \cite{WangPei2018-SecurityAnalysis} and  DLV \cite{HuangKwiatkowska2017} combine reachability with search; Sherlock \cite{DuttaJha2018}, Reluplex \cite{KatzBarrett2017}, Planet \cite{Ehlers2017} and BaB \cite{BunelTurkaslan2018} combine search with optimization. Lie et al. \cite{LiuArnon2019} provide a more detailed comparison and discussion of the above mentioned work.
\section{Conclusion and Future Work}\label{sec:Conclusion}
We proposed and implemented \tool, a novel blackbox-fuzzing technique
for attacking deep neural networks. \tool is simple and effective in
finding adversarial examples with low distortion, and it outperforms state-of-the-art blackbox attacks in a query-limited
setting. In our future work, we will continue improving the effectiveness of \tool for an even more query-efficient $L_{\infty}$ attack. We are also interested in extending \tool to construct query-efficient $L_{2}$ attacks.

Designing effective defenses to secure deep neural networks against
adversarial attacks is non-trivial. In this paper, we did not focus on
proposing defenses against blackbox attacks. Instead, we attacked
neural networks with adversarial training-based
defenses \cite{MadryMakelov2018}. Another interesting direction for
future work is to develop defense techniques that specifically target
blackbox attacks, for instance by identifying
patterns in their sequences of queries.

\bibliographystyle{ACM-Reference-Format}
\bibliography{reference}

\balance
\newpage
\section{Appendix}
\noindent{\bf Theorem} \ref{theorem-maximum-minimum}
Given a linear classification function
$f(\textbf{x})=\textbf{w}^{T}\textbf{x}+b$, where
$\textbf{w}^{T}=(w_{1},...,w_{n})$ and $b\in\mathbb{R}$, an
$n$-dimensional cube $\mathcal{I}=I_{1}\times...\times I_{n}$, where
$I_{i}=[l_{i},u_{i}]$ for $1\leq i\leq n$, and an input
$\textbf{x}\in \mathcal{I}$, we have:
\begin{enumerate}
	\item $\min$ $f(\mathcal{I}) = f(\textbf{x}')$, where
	$\textbf{x}'(i)=l_{i}$ (resp. $\textbf{x}'(i)=u_{i}$) if
	$f(\textbf{x}[u_{i}/x_{i}])> f(\textbf{x}[l_{i}/x_{i}])$
	(resp. $f(\textbf{x}[u_{i}/x_{i}])\leq
	f(\textbf{x}[l_{i}/x_{i}])$) for $1\leq i\leq
	n$
	\item $\max$ $f(\mathcal{I}) =
	f(\textbf{x}')$, where $\textbf{x}'(i)=u_{i}$
	(resp. $\textbf{x}'(i)=l_{i}$) if $f(\textbf{x}[u_{i}/x_{i}])>
	f(\textbf{x}[l_{i}/x_{i}])$
	(resp. $f(\textbf{x}[u_{i}/x_{i}])\leq
	f(\textbf{x}[l_{i}/x_{i}])$) for $1\leq i\leq n$
\end{enumerate}
\begin{proof}
	This theorem can be proved by induction on the number of dimensions of \textbf{x}. We only show the proof for case $(1)$, as case $(2)$ can be proved similarly.
	
	\textbf{Base case $n=1$}:  In the one dimensional  space, let $\textbf{x}=x_{1}$,  $f(\textbf{x})=w_{1}x_{1}+b$ and $\mathcal{I}= I_{1}=[l_{1},u_{1}]$.  We have three cases as follows: $w_{1}>0$, $w_{1}<0$ and $w_{1}=0$.
	
	Assume that $w_{1}>0$. We have that $\min\ f(\mathcal{I}) = \min\ \{f(\textbf{x})\ |\ \textbf{x}\in\mathcal{I}\} = \min\ \{ w_{1}x_{1}+b\ |\ x_{1}\in I_{1} \}=w_{1}l_{1}+b$. Let $\textbf{x}'=x_{1}'$ be an input, where $\textbf{x}'(1)=l_{1}$ (resp. $\textbf{x}'(1)=u_{1}$) if $f(\textbf{x}[u_{1}/x_{1}])> f(\textbf{x}[l_{1}/x_{1}])$
	(resp. $f(\textbf{x}[u_{1}/x_{1}])\leq f(\textbf{x}[l_{1}/x_{1}])$). Since $f(\textbf{x}[u_{1}/x_{1}]) = f(u_{1})  >  f(l_{1}) = f(\textbf{x}[l_{1}/x_{1}])$, we have that $\textbf{x}'(1)=l_{1}$. This means $f(\textbf{x}') = w_{1}l_{1}+b$. Therefore, we have that  $\min\ f(\mathcal{I}) = f(\textbf{x}')$.
	
	The case for $w_{1}<0$ can be proved similarly.
	
	Assume that $w_{1}=0$. We have that $\min\ f(\mathcal{I}) = \min\ \{f(\textbf{x})\ |\ \textbf{x}\in\mathcal{I}\} = \min\ \{ 0x_{1}+b\ |\ l_{1}\leq x_{1}\leq u_{1} \}=b$. Since $f(\textbf{x}') = 0\textbf{x}'(1)+b =b$ for any input $\textbf{x}'$, we have that  $\min\ f(\mathcal{I}) = f(\textbf{x}')$, where $\textbf{x}'(1)=l_{1}$ (resp. $\textbf{x}'(1)=u_{1}$) if $f(\textbf{x}[u_{1}/x_{1}])> f(\textbf{x}[l_{1}/x_{1}])$ (resp. $f(\textbf{x}[u_{1}/x_{1}])\leq f(\textbf{x}[l_{1}/x_{1}]))$.
	
	\textbf{Induction step}: Assume that case $(1)$ holds for $n\leq k$. We now consider the case $n=k+1$. In $k+1$ dimensional space, let $\textbf{x}=(x_{1},...,x_{k},x_{k+1})^{T}$, $\mathcal{I}= I_{1}\times...\times I_{k}\times I_{k+1}$ and $f(\textbf{x})=\sum_{i=1}^{k+1}w_{i}x_{i}+b$. In the following, we write $\textbf{x}_{1..k}$ to mean $(x_{1},...,x_{k})^{T}$ and write $\textbf{x}_{k+1..k+1}$ to mean $x_{k+1}$. Let $f_{1}(\textbf{x}_{1..k})=\sum_{i=1}^{k}w_{i}x_{i}+b$ and $f_{2}(\textbf{x}_{k+1..k+1}) = w_{k+1}x_{k+1}$ be linear functions. Hence, we have that $f(\textbf{x})=\sum_{i=1}^{k+1}w_{i}x_{i}+b = \sum_{i=1}^{k}w_{i}x_{i}+b + w_{k+1}x_{k+1} = f_{1}(\textbf{x}_{1..k}) + f_{2}(\textbf{x}_{k+1..k+1})$. 
	
	First, we have that $\min\ f(\mathcal{I}) = \min\ \{f(\textbf{x})\ |\ \textbf{x}\in\mathcal{I}\} = \min$ $\{f_{1}(\textbf{x}_{1..k}) + f_{2}(\textbf{x}_{k+1..k+1})\ |\ l_{i}\leq x_{i}\leq u_{i},$ where $1\leq i\leq k+1\}= \min$ $\{f_{1}(\textbf{x}_{1..k}) \ |\ \textbf{x}_{1..k}\in I_{1}\times...\times I_{k}\} + \min$ $\{\ f_{2}(\textbf{x}_{k+1..k+1})\ |\ x_{k+1}\in I_{k+1}\}$. Let $\textbf{x}'=(x_{1}',...,x_{k}',x_{k+1}')^{T}$ be an input, where $\textbf{x}'(i)=l_{i}$ (resp. $\textbf{x}'(i)=u_{i}$) if $f(\textbf{x}[u_{i}/x_{i}])> f(\textbf{x}[l_{i}/x_{i}])$ (resp. $f(\textbf{x}[u_{i}/x_{i}])\leq f(\textbf{x}[l_{i}/x_{i}])$) for $1\leq i\leq k+1$. Let $\textbf{x}'_{1..k}$ denote $(x_{1}',...,x_{k}')^{T}$ and $\textbf{x}'_{k+1..k+1}$ denote $x_{k+1}'$. From the induction hypothesis, we know that $\min$ $\{f_{1}(\textbf{x}_{1..k}) \ |\ l_{i}\leq x_{i}\leq u_{i},$ where $1\leq i\leq k\} = f_{1}(\textbf{x}'_{1..k})$ and $\min\ \{\ f_{2}(\textbf{x}_{k+1..k+1})\ |\ x_{k+1}\in I_{k+1}\}=f_{2}(\textbf{x}'_{k+1..k+1})$. Therefore, we have that $\min\ f(\mathcal{I}) = f_{1}(\textbf{x}'_{1..k}) + f_{2}(\textbf{x}'_{k+1..k+1})=f(\textbf{x}')$
	
	From above, we know that case $(1)$ holds for all $n\geq 1$.  Since case  $(2)$ can be proved similarly, we know that Theorem \ref{theorem-maximum-minimum} holds for all $n\geq 1$. This proves Theorem \ref{theorem-maximum-minimum}.
\end{proof}

\noindent{\bf Theorem} \ref{theorem-projection}
Let $f(\textbf{x})=\textbf{w}^{T}\textbf{x}+b$ be a linear
classification function, and $\mathcal{I}_{1}$, $\mathcal{I}_{2}$ two
$n$-dimensional cubes such that
$\mathcal{I}_{1}\subseteq\mathcal{I}_{2}$. Assuming that $\textbf{x}$
is a vertex of
$\mathcal{I}_{2}$, we have:
\begin{enumerate}
	\item if $\min$  $f(\mathcal{I}_{2})=f(\textbf{x})$, then $\min$ $f(\mathcal{I}_{1}) = f(\textsc{Proj}(\mathcal{I}_{1},\textbf{x}))$
	\item if $\max$ $f(\mathcal{I}_{2})=f(\textbf{x})$, then $\max$ $f(\mathcal{I}_{1}) = f(\textsc{Proj}(\mathcal{I}_{1},\textbf{x}))$
\end{enumerate}
\begin{proof}
	This theorem can be proved by induction on the number of dimensions of \textbf{x}. We only show the proof for case $(1)$, as case $(2)$ can be proved similarly.
	
	\textbf{Base case $n=1$}:  In this case, let $\textbf{x}=x_{1}$ and $f(\textbf{x})=w_{1}x_{1}+b$. Let $\mathcal{I}_{1}= [l_{1},u_{1}]$ and $\mathcal{I}_{2}= [l_{2},u_{2}]$ be two one-dimensional cubes such that $l_{2}\leq l_{1}$ and $u_{1}\leq u_{2}$. Assume that $\min$  $f(\mathcal{I}_{2})=f(\textbf{x})$. We have three cases as follows: $w_{1}>0$, $w_{1}<0$ and $w_{1}=0$. 
	
	Assume that $w_{1}>0$. From Theorem $1$, we know that $\min\ f(\mathcal{I}_{2}) = f(l_{2})$. Since $f$ is an injective function, we have that $\textbf{x}=l_{2}$. From Theorem $1$, we also know that $\min\ f(\mathcal{I}_{1})=  f(l_{1})$. Since $l_{2}\leq l_{1}$, we have that $\textsc{Proj}(\mathcal{I}_{1},\textbf{x}) = \textsc{Proj}(\mathcal{I}_{1},l_{2}) = l_{1}$.
	Therefore, we have $\min\ f(\mathcal{I}_{1})=f(\textsc{Proj}(\mathcal{I}_{1},\textbf{x}))$.
	
	The proof for case $w_{1}<0$ is similar. 
	
	Assume that $w_{1}=0$. Since $f(\textbf{x}') = b$ for any $\textbf{x}'$, it is trivially true that $\min\ f(\mathcal{I}_{1})= b =f(\textsc{Proj}(\mathcal{I}_{1},\textbf{x}))$.
	
	\textbf{Induction step}: Assume that case $(1)$ holds for $n\leq k$. We now consider the case $n=k+1$. In $k+1$ dimensional space, let $\textbf{x}=(x_{1},...,x_{k},x_{k+1})^{T}$ and $f(\textbf{x})=\sum_{i=1}^{k+1}w_{i}x_{i}+b$. In the following, we write $\textbf{x}_{1..k}$ to mean $(x_{1},...,x_{k})^{T}$ and write $\textbf{x}_{k+1..k+1}$ to mean $x_{k+1}$. Let $f_{1}(\textbf{x}_{1..k})=\sum_{i=1}^{k}w_{i}x_{i}+b$ and $f_{2}(\textbf{x}_{k+1..k+1}) = w_{k+1}x_{k+1}$ be linear functions. Hence, we have $f(\textbf{x})=\sum_{i=1}^{k+1}w_{i}x_{i}+b = \sum_{i=1}^{k}w_{i}x_{i}+b + w_{k+1}x_{k+1} = f_{1}(\textbf{x}_{1..k}) + f_{2}(\textbf{x}_{k+1..k+1})$. 
	Let $\mathcal{I}_{1}$ and $\mathcal{I}_{2}$ be two $k+1$-dimensional cubes such that $\mathcal{I}_{1}\subseteq\mathcal{I}_{2}$.
	For an $k+1$ dimensional cube $\mathcal{I}= I_{1}\times...\times I_{k}\times I_{k+1}$, we write $\mathcal{I}^{1..k}$ to denote $I_{1}\times...\times I_{k}$ and write $\mathcal{I}^{k+1..k+1}$ to denote $I_{k+1}$.
	
	Notice that the most interesting case is when $w_{i}\neq0$ for all $1\leq i\leq k+1$ and all other cases are simpler. When $w_{i}=0$ for some $1\leq i\leq k+1$, case $(1)$ can be reduced into an equivalent case in $m$-dimensional space, where $m\leq k$, and we can prove the equivalent case using induction hypothesis directly. In the following, we prove the case when $w_{i}\neq0$ for all $1\leq i\leq k+1$.
	
	Assume that $\min\ f(\mathcal{I}_{2})=f(\textbf{x}) = f_{1}(\textbf{x}_{1..k}) + f_{2}(\textbf{x}_{k+1..k+1})$. Since  \textbf{x} is a vertex on $\mathcal{I}_{2}$ and $f$ is an injective function, according to Theorem $1$, we know that $\textbf{x}(i)=l_{i}$ (resp. $\textbf{x}(i)=u_{i}$) if $f(\textbf{x}''[u_{i}/x_{i}])> f(\textbf{x}''[l_{i}/x_{i}])$ (resp. $f(\textbf{x}''[u_{i}/x_{i}])\leq f(\textbf{x}''[l_{i}/x_{i}])$) for $1\leq i\leq k+1$, where $\textbf{x}''$ is an arbitrary input on $\mathcal{I}_{2}$. According to Theorem $1$, we also have that $\min\ f_{1}(\mathcal{I}_{2}^{1..k}) = f_{1}(\textbf{x}_{1..k})$ and $\min\ f_{2}(\mathcal{I}_{2}^{k+1..k+1}) = f_{2}(\textbf{x}_{k+1..k+1})$. From the induction hypothesis and above, we have that $\min\ f_{1}(\mathcal{I}_{1}^{1..k}) = f_{1}(\textsc{Proj}(\mathcal{I}_{1}^{1..k},\textbf{x}_{1..k}))$ and that $\min\ f_{2}(\mathcal{I}_{1}^{k+1..k+1}) = f_{2}(\textsc{Proj}(\mathcal{I}_{1}^{k+1..k+1},\textbf{x}_{k+1..k+1}))$.
	
	From Theorem $1$, we have that $\min\ f(\mathcal{I}_{1})=f(\textbf{x}') = f_{1}(\textbf{x}'_{1..k}) + f_{2}(\textbf{x}_{k+1..k+1}')$, where $\textbf{x}'$ is a vertex on $\mathcal{I}_{1}$ such that $\textbf{x}'(i)=l_{i}$ (resp. $\textbf{x}'(i)=u_{i}$) if $f(\textbf{x}''[u_{i}/x_{i}])> f(\textbf{x}''[l_{i}/x_{i}])$ (resp. $f(\textbf{x}''[u_{i}/x_{i}])\leq f(\textbf{x}''[l_{i}/x_{i}])$) for $1\leq i\leq k+1$ and an arbitrary input $\textbf{x}''$ on $\mathcal{I}_{1}$. Also, according to Theorem $1$, we have that $\min\ f_{1}(\mathcal{I}_{1}^{1..k}) = f_{1}(\textbf{x}'_{1..k})$ and $\min\ f_{2}(\mathcal{I}_{1}^{k+1..k+1}) = f_{2}(\textbf{x}'_{k+1..k+1})$. Therefore, we have that $\min\ f(\mathcal{I}_{1}) = f_{1}(\textsc{Proj}(\mathcal{I}_{1}^{1..k},\textbf{x}_{1..k})) + f_{2}(\textsc{Proj}(\mathcal{I}_{1}^{k+1..k+1},\textbf{x}_{k+1..k+1}))$.
	
	From the definition of the projection operator $\textsc{Proj}$, we have that $f(\textsc{Proj}(\mathcal{I}_{1},\textbf{x})) = f_{1}(\textsc{Proj}(\mathcal{I}_{1}^{1..k},\textbf{x}_{1..k})) + f_{2}(\textsc{Proj}(\mathcal{I}_{1}^{k+1..k+1},\newline\textbf{x}_{k+1..k+1}))$. Therefore, we have that   $\min\ f(\mathcal{I}_{1}) = f(\textsc{Proj}(\mathcal{I}_{1},\textbf{x}))$.
	
	From above, we know that case $(1)$ holds for all $n\geq 1$.  Since case  $(2)$ can be proved similarly, we know that Theorem \ref{theorem-projection} holds for all $n\geq 1$. This proves Theorem \ref{theorem-projection}.
\end{proof}

\end{document}